\documentclass{article}
\usepackage{todonotes}
\usepackage[a4paper,margin=1in]{geometry}
\usepackage{amsmath,amssymb,amsthm,bm,mathtools}
\usepackage{graphicx}
\graphicspath{{./figures/}}
\usepackage{xcolor}
\usepackage{booktabs}
\usepackage[table]{xcolor}
\usepackage{multirow}
\usepackage{caption}
\usepackage{subcaption}
\usepackage{authblk}
\usepackage{cite}
\usepackage[ruled,linesnumbered]{algorithm2e}
\usepackage{tikz}
\usetikzlibrary{arrows.meta,positioning,fit,calc,decorations.pathmorphing,shapes.geometric}
\usepackage[colorlinks=true,linkcolor=blue,citecolor=blue,urlcolor=blue]{hyperref}

\newcommand{\R}{\mathbb{R}}

\newcommand{\I}{\mathbb{I}}

\newcommand{\cG}{\mathcal{G}}
\newcommand{\cP}{\mathcal{P}}
\newcommand{\cB}{\mathcal{B}}
\newcommand{\cR}{\mathcal{R}}
\newcommand{\cT}{\mathcal{T}}
\newcommand{\cS}{\mathcal{S}}
\newcommand{\cK}{\mathcal{K}}

\newcommand{\cE}{\mathcal{E}}

\newcommand{\diff}{\mathop{}\!\mathrm{d}}
\newcommand{\inner}[2]{\left\langle #1,#2\right\rangle}
\newcommand{\norm}[1]{\left\lVert #1\right\rVert}
\newcommand{\softmax}{\operatorname{softmax}}
\newcommand{\ELU}{\operatorname{ELU}}

\begin{document}
\title{Let There Be Light: Reflection, Refraction and Scattering for Neural Operators}

\author[1,2,4]{Keke Wu}
\author[5]{Yixuan Zhang}
\author[1,3,4 \thanks{Corresponding author:jingrunchen@ustc.edu.cn}]{Jingrun Chen}
\affil[1]{\small Suzhou Institute for Advanced Research, University of Science and Technology of China, Suzhou 215123, China}
\affil[2]{\small School of Artificial Intelligence and Data Science, University of Science and Technology of China, Hefei 230026, China }
\affil[3]{\small School of Mathematical Sciences, University of Science and Technology of China, Hefei 230026, China}
\affil[4]{\small Suzhou Big Data \& AI Research and Engineering Center, Suzhou 215123, China}
\affil[5]{\small School of Mathematical Science, Peking University, Beijing, 100871, China}

\date{\today}

\maketitle

\begin{abstract}
Neural operators learn mappings between infinite-dimensional function spaces and provide a data-driven surrogate modeling paradigm for parametric partial differential equations (PDEs). Existing architectures typically obtain expressivity by parameterizing integral kernels in prescribed transform domains or by applying attention-like interactions over discretized spatial points. 
While these approaches have achieved substantial progress, they often face a persistent trade-off among physical interpretability, nonlocal spatial communication, mesh scalability, and computational cost. We propose a \emph{Light-inspired neural operator} (LiNO), an operator-learning architecture whose latent evolution is decomposed into three mechanisms motivated by elementary light transport: reflection, refraction, and scattering. Reflection and refraction act as adaptive pointwise transformations in latent feature space, enabling local feature reorientation and anisotropic modulation, whereas scattering performs input-dependent nonlocal propagation over the physical domain. We first formulate scattering as a normalized pairwise kernel with relative positional bias, and then develop an efficient scattering variant that replaces explicit pairwise interactions with positive-feature global propagation and a local diffusion branch, reducing the dominant spatial complexity from quadratic to linear. This yields a structured neural operator that separates local feature modulation from global spatial communication while retaining a modular and interpretable latent evolution. Experiments on Burgers' equation, Darcy flow, transonic airfoil flow, and Navier--Stokes dynamics show that LiNO achieves competitive accuracy across representative benchmarks, improves over reported values on several settings, scales effectively to high-resolution grids, captures geometry-dependent flow fields, and maintains stable autoregressive rollouts for dynamic problems.
These results suggest that light-inspired latent transport provides a flexible design principle for building interpretable and scalable neural operators for PDE surrogate modeling. 
\end{abstract}

\section{Introduction}

Parametric partial differential equations (PDEs) commonly model complex dynamics in inverse design, uncertainty quantification, and optimal control. Efficiently solving such problems repeatedly poses a major computational bottleneck in scientific computing. Abstractly, given a family of PDEs indexed by input functions, material coefficients, initial/boundary data, source terms, or geometric descriptors, the central mathematical object is the solution operator that maps each parameter instance to the corresponding solution
\begin{equation}
    \mathcal{G}: \mathcal{A} \to \mathcal{U}, \qquad a \mapsto u=\mathcal{G}(a),
\end{equation}
where $\mathcal{A}$ and $\mathcal{U}$ are infinite-dimensional function spaces. Classical numerical methods, such as finite difference, finite volume, spectral, and finite element methods, approximate each new instance by discretizing the governing equation and solving the resulting algebraic system. These solvers remain indispensable due to their consistency, stability, convergence theory, and mature treatment of complex physics. However, their repeated deployment can be prohibitively expensive when many-query evaluations are required, for example, in Bayesian inversion, design-space exploration, ensemble forecasting, and real-time control. Neural operators provide an alternative surrogate modeling paradigm: rather than learning a finite-dimensional input--output map tied to a fixed discretization, they aim to learn the operator $\mathcal{G}$ itself from data and evaluate it rapidly on unseen input functions, ideally with a degree of discretization invariance and resolution transfer.

Operator learning has rapidly evolved from early universal-approximation formulations to scalable architectures for PDE-governed systems. DeepONet introduced a branch--trunk decomposition for nonlinear operator approximation and connected neural-network architectures with universal approximation theorems for operators \cite{chen1995,lu2021deeponet}. The general neural-operator framework of Kovachki et al. further formalized operator learning as compositions of nonlinear activations and learnable integral operators, leading to graph, low-rank, multipole, and Fourier parameterizations \cite{kovachki2023neuraloperator}. Among these models, the Fourier neural operator (FNO) parameterizes translation-invariant integral kernels in Fourier space and has become a canonical baseline for parametric PDEs such as Burgers' equation, Darcy flow, and Navier--Stokes equations \cite{li2021fno}. In parallel, graph kernel neural operators and multipole graph neural operators introduced graph-based discretization-invariant kernel approximations and multilevel interaction mechanisms for irregular data structures \cite{li2020gno,li2020multipole}. Systematic benchmark studies have shown that DeepONet and FNO exhibit complementary advantages: FNO is highly efficient on regular grids, while DeepONet-type formulations can be more flexible in input--output function representations and boundary-condition handling \cite{lu2022comprehensive}. 

Subsequent developments have expanded neural operators along several important directions. Transform-domain neural operators replace or augment Fourier representations with alternative bases or transforms. The Laplace neural operator (LNO) exploits pole--residue representations in the Laplace domain and is particularly motivated by non-periodic signals and transient responses \cite{cao2024laplace}. Wavelet neural operators (WNOs) use localized time--frequency representations to better capture multiscale and localized spatial features \cite{tripura2023wavelet}. Architecture-level extensions, including U-FNO and U-shaped neural operators, improve the expressivity and memory efficiency of Fourier-type layers by incorporating multiresolution encoder--decoder structures and local convolutional paths \cite{wen2022ufno,rahman2022uno}. Geometry-aware variants, including Geo-FNO, GINO, and point-cloud neural operators, address the restriction of FFT-based models to rectangular uniform grids by introducing learned deformations, signed-distance or point-cloud geometric encodings, graph-Fourier hybrids, or continuous point-cloud formulations \cite{li2023geofno,li2023gino,zeng2025pcno,han2026geometric}. Dynamical-systems-inspired formulations, such as the Koopman neural operator, seek to improve long-time prediction by learning latent representations in which nonlinear PDE dynamics can be advanced through approximately linear evolution \cite{xiong2024koopman,xiong2023koopmanlab}.

Attention-based operator learning has also become increasingly prominent. Since self-attention provides a data-dependent mechanism for modeling long-range interactions \cite{vaswani2017attention}, it has been adapted to PDE solution operators through Galerkin-type transformer architectures. The Galerkin Transformer shows that softmax normalization is not essential for operator learning and connects linear attention with Petrov--Galerkin projection viewpoints \cite{cao2021galerkin}. OFormer further develops a transformer-based encoder--decoder architecture for PDE operator learning, where the input function is encoded into latent spatial representations and decoded at arbitrary query locations through attention-based interactions \cite{li2023oformer}.
Physics-aware and geometry-aware transformer solvers further exploit attention to organize information over irregular meshes and latent physical states. Transolver introduces Physics-Attention, which adaptively partitions mesh points into learnable slices associated with latent physical states and performs attention over the resulting physics-aware tokens, thereby enabling transformer-based PDE solving on general geometries
\cite{wu2024transolver}. Transolver++ further scales this paradigm to million-scale geometries through a highly parallel implementation and a local adaptive mechanism for learning more expressive physical-state representations \cite{luo2025transolverpp}. More recently, Transolver-3 targets industrial-scale geometries with over $10^8$ cells by introducing faster slice/deslice operations, geometry slice tiling, amortized training on high-resolution meshes,
and physical-state caching during inference \cite{zhou2026transolver3}.
These models suggest that data-dependent nonlocal interaction is crucial for complex PDEs, but they also highlight persistent challenges: standard attention over $N$ spatial degrees of freedom has $\mathcal{O}(N^2)$ memory and computational complexity, while purely spectral kernels may impose restrictive inductive biases and can struggle with non-periodicity, localized structures, sharp interfaces, or complex geometries.

Despite this progress, there remains a need for neural-operator architectures that combine three desirable properties: physically interpretable latent evolution, efficient nonlocal interaction, and flexibility beyond fixed Fourier kernels. Many existing methods can be interpreted as choosing a particular mechanism for information propagation: spectral convolution transports information globally through Fourier modes; graph neural operators propagate information through learned kernel interactions on nodes; transformers exchange information through attention; and Koopman-type methods evolve latent observables through approximately linear dynamics. In this work, we propose a complementary viewpoint inspired by light transport. We regard the latent feature field as a collection of light-like feature rays distributed over the computational domain. Its evolution is decomposed into three physically motivated mechanisms: \emph{reflection}, which locally redirects latent features in feature space; \emph{refraction}, which adaptively modulates feature transmission according to state-dependent anisotropy; and \emph{scattering}, which couples spatially separated locations through a learnable interaction kernel. This decomposition leads to a Light-inspired Neural Operator (LiNO) whose basic block combines local feature-space transformations with global or efficient spatial scattering.

The proposed light-inspired formulation is designed to bridge physical interpretability and computational efficiency. Reflection and refraction act pointwise on the latent channel dimension and therefore introduce nonlinear local feature evolution at negligible spatial cost. Scattering acts over the physical domain and supplies the nonlocal communication required by elliptic, parabolic, and fluid-type PDE solution operators. We first formulate a full scattering layer as a relative-position-biased attention kernel over spatial nodes, which provides a direct and expressive discretization of nonlocal light scattering. To reduce the quadratic cost of this formulation, we further develop an efficient scattering layer based on linearized attention together with a local depthwise convolutional branch, preserving global information exchange while lowering the dominant complexity from quadratic to linear in terms of the number of spatial nodes. The resulting architecture retains the operator-learning perspective of existing neural operators while introducing a modular optical inductive bias that separates local feature redirection, anisotropic transmission, and nonlocal spatial mixing.

\begin{figure}[htbp]
    \centering
    \includegraphics[width=0.8\linewidth]{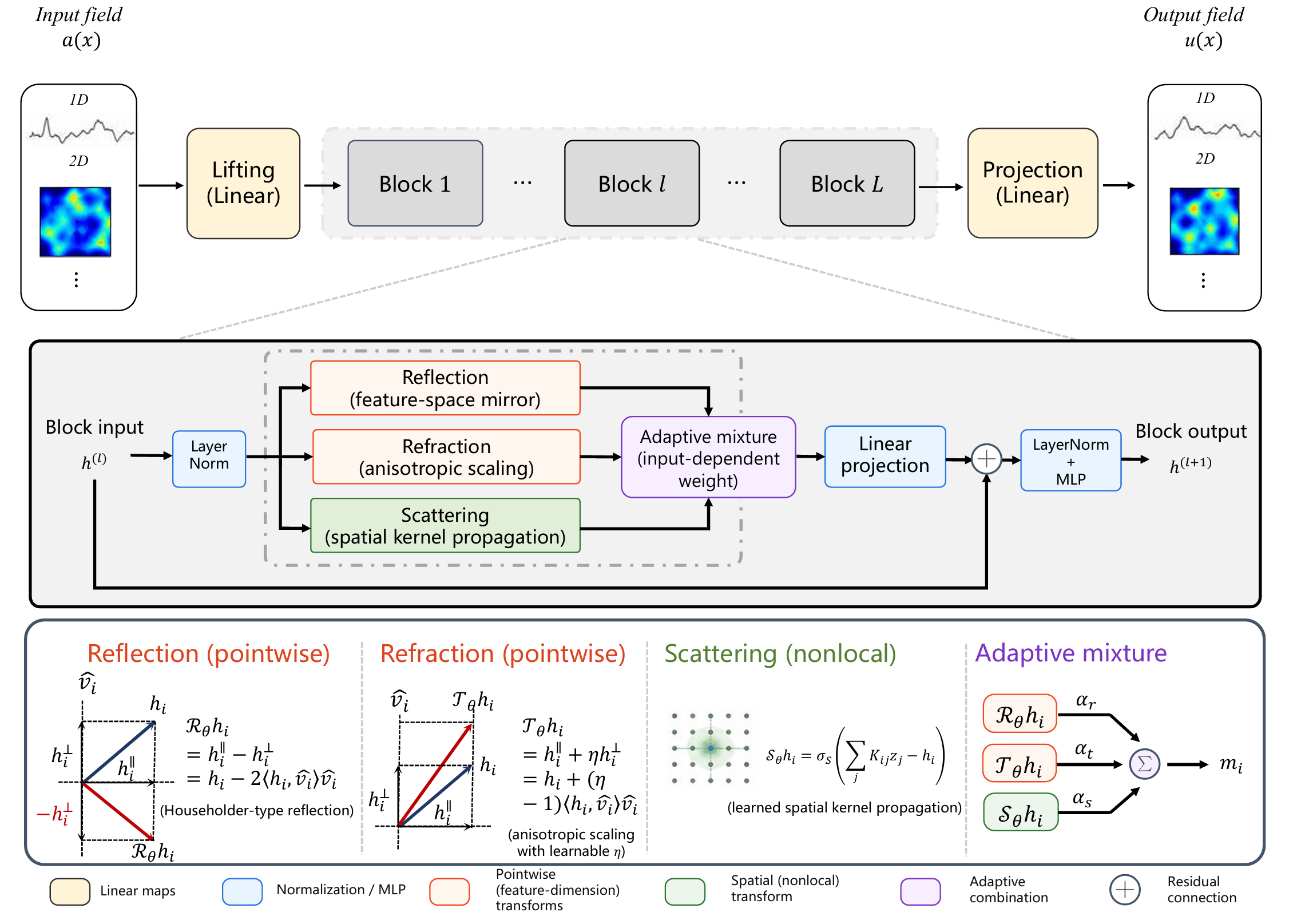}
    \caption{
    \textbf{Architecture of the Light-inspired neural operator.}
    \textbf{Top:} LiNO learns the parametric PDE solution operator 
    $\mathcal{G}:a\mapsto u$ by lifting the input field to a latent feature space, 
    evolving it through stacked light-evolution blocks, and projecting the final latent 
    representation to the output field. 
    \textbf{Middle:} Each light-evolution block updates the latent state through three 
    parallel branches. 
    Reflection and refraction are pointwise feature-space transformations, while scattering 
    performs spatial kernel propagation. 
    Their outputs are combined by an input-dependent adaptive mixture and inserted into a 
    residual block with linear projection and pointwise MLP correction. 
    \textbf{Bottom:} The elementary operations underlying the block. 
    Reflection acts as an adaptive Householder mirror in feature space; refraction rescales 
    the component along a learned normal direction while preserving the tangential component; 
    scattering realizes nonlocal propagation by a learned spatial kernel; and adaptive 
    mixture weights determine the relative contribution of the three light-inspired 
    mechanisms. 
    }
    \label{fig:lino}
\end{figure}

The main contributions of this work are summarized as follows. First, we introduce a light-inspired operator-learning architecture for parametric PDEs, in which the latent evolution is explicitly decomposed into reflection, refraction, and scattering mechanisms. Second, we formulate the reflection and refraction layers as adaptive feature-space transformations motivated by geometric optics, while the scattering layer is derived as a spatial nonlocal kernel with relative positional bias. Third, we propose an efficient scattering variant that combines linear attention with local convolution, substantially reducing the computational burden of full spatial attention. Finally, we demonstrate through canonical operator-learning benchmarks, including one-dimensional Burgers' equation, two-dimensional Darcy flow, and time-dependent Navier--Stokes dynamics, that the proposed model provides a flexible and interpretable alternative to purely Fourier-, graph-, or transformer-based neural operators. A schematic of the proposed LiNO is presented in Fig.~\ref{fig:lino}.

The remainder of this paper is organized as follows. 
Section~\ref{sec:method} presents the proposed Light-inspired Neural Operator. 
We first formulate the operator-learning problem and then introduce the light-evolution block, whose latent update is decomposed into reflection, refraction, and scattering. 
Reflection and refraction are developed as adaptive feature-space transformations, while scattering is formulated as a nonlinear spatial propagation operator. 
We further develop a scalable scattering implementation that replaces explicit pairwise interactions with positive-feature global propagation and a local diffusion branch, reducing the dominant spatial complexity from quadratic to linear.
Section~\ref{sec:experiments} reports numerical experiments on Burgers' equation, Darcy flow, transonic airfoil flow, and two-dimensional Navier--Stokes dynamics, covering accuracy, computational efficiency, component ablations, geometry-dependent prediction, and autoregressive stability. 
Section~\ref{sec:conclusion} summarizes the main findings and outlines future directions. 
Appendix~\ref{app:benchmark-pdes} provides the benchmark PDEs, dataset details, and implementation settings, while Appendix~\ref{app:exp-kernel-feature} discusses the exponential-kernel feature representation used to motivate efficient scattering.

\section{Method}
\label{sec:method}
\subsection{Problem setting and neural-operator formulation}

Let $\{x_i\}_{i=1}^{N}\subset\Omega$ be a discretization of the spatial domain. 
In one dimension, $N=Q$ for a grid with $Q$ points; in two dimensions, 
$N=HW$ for a Cartesian grid of size $H\times W$. 
The discretized input field is represented as
\begin{equation}
    a_N = \{a(x_i)\}_{i=1}^{N} \in \R^{N\times d_{\mathrm{in}}},
\end{equation}
where the first dimension indexes spatial locations and the second dimension indexes input channels. 

The proposed LiNO takes the compositional form
\begin{equation}
    \cG_\theta
    =
    \Pi_\theta \circ
    \left(\cB_\theta^{(L)}\circ\cdots\circ\cB_\theta^{(1)}\right)
    \circ \cP_\theta,
    \label{eq:global-lightno}
\end{equation}
where $\cP_\theta$ is a lifting map, $\Pi_\theta$ is a projection map, and $\cB_\theta^{(\ell)}$ is the $\ell$-th light-evolution block. The latent field at layer $\ell$ is denoted by
\begin{equation}
    h^{(\ell)}:\Omega\to\R^M, \qquad
    h_i^{(\ell)} \approx h^{(\ell)}(x_i)\in\R^M,
\end{equation}
where $M$ denotes the latent feature dimension.
In the optical analogy, $h_i^{(\ell)}$ is the latent optical state carried by the spatial location $x_i$.

The lifting and projection are pointwise affine maps,
\begin{align}
    h_i^{(0)} &= \cP_\theta(a)(x_i)=W_{\mathrm{in}} a(x_i)+b_{\mathrm{in}},
    \label{eq:lifting}\\
    \widehat{u}_i &= \Pi_\theta(h^{(L)})(x_i)=W_{\mathrm{out}}h_i^{(L)}+b_{\mathrm{out}}.
    \label{eq:projection}
\end{align}
Coordinate channels may be concatenated with the physical input field before lifting, so that the operator has access to explicit spatial-position information. In that case, $a_i$ denotes the augmented input feature at $x_i$, including both physical variables and coordinate features.

\subsection{Light-evolution block: reflection, refraction, and scattering}

Each light-evolution block updates the latent field through a residual rule
\begin{equation}
    h^{(\ell+1)} = h^{(\ell)} + \cE_\theta^{(\ell)}(h^{(\ell)}),
    \label{eq:residual-evolution}
\end{equation}
where $\cE_\theta^{(\ell)}$ is decomposed into three light-inspired components:
\begin{equation}
    \cE_\theta^{(\ell)}
    =
    \alpha_r^{(\ell)}\cR_\theta^{(\ell)}
    +
    \alpha_t^{(\ell)}\cT_\theta^{(\ell)}
    +
    \alpha_s^{(\ell)}\cS_\theta^{(\ell)}.
    \label{eq:light-decomposition}
\end{equation}
Here $\cR_\theta$ is reflection, $\cT_\theta$ is refraction, and $\cS_\theta$ is scattering. The reflection and refraction maps act independently at each spatial location and only mix the latent feature dimension. The scattering map operates over the physical domain and is responsible for the propagation of spatial information.

\subsection{Reflection: adaptive feature-space mirror transformation}

In physical optics, reflection changes the direction of propagation according to the local interface normal. In LiNO, we reinterpret this mechanism in feature space: the latent vector $h_i\in\R^M$ at each spatial point is reflected with respect to an adaptive hyperplane whose normal direction is learned from the current state.

For each spatial index $i$, define an adaptive direction
\begin{equation}
    v_i = W_r h_i + b_r, \qquad
    \widehat{v}_i = \frac{v_i}{\norm{v_i}_2},
    \label{eq:adaptive-normal}
\end{equation}
where $W_r\in\R^{M\times M}$ and $b_r\in\R^M$ are trainable parameters. The reflection operator is given by the Householder-type formula
\begin{equation}
    \cR_\theta(h_i)
    =
    h_i - 2\inner{h_i}{\widehat{v}_i}\widehat{v}_i.
    \label{eq:reflection}
\end{equation}
For a fixed unit direction $\widehat v_i$, this transformation is equivalent to applying the Householder matrix
\begin{equation}
    H_i = I_M - 2\widehat v_i \widehat v_i^\top
\end{equation}
to the feature vector $h_i$, namely
\begin{equation}
    \cR_\theta(h_i)=H_i h_i.
\end{equation}
A direct implementation of this matrix-vector product would require forming the dense matrix
$H_i\in\R^{M\times M}$, leading to $O(M^2)$ storage and computational cost per spatial location. In contrast, we use the matrix-free Householder identity in \eqref{eq:reflection}: the reflection is computed only through one inner product
$\inner{h_i}{\widehat v_i}$ and one rank-one rescaling of $\widehat v_i$. Consequently, the cost of the reflection step is reduced to $O(M)$ per spatial point, while retaining the exact action of the corresponding Householder reflection.

Geometrically, \eqref{eq:reflection} reverses the component of $h_i$ along $\widehat v_i$ while keeping the tangential component unchanged. Indeed, writing
\begin{equation}
    h_i^{\perp}=\inner{h_i}{\widehat v_i}\widehat v_i,
    \qquad
    h_i^{\parallel}=h_i-h_i^{\perp},
\end{equation}
we obtain
\begin{equation}
    \cR_\theta(h_i)=h_i^{\parallel}-h_i^{\perp}.
\end{equation}
Thus, reflection provides a learnable, direction-dependent redistribution of latent features without introducing cross-site spatial coupling. This is useful when the input field induces sharp changes, interfaces, or local directional modes, because the model can locally reorient the latent representation before applying spatial propagation.

\subsection{Refraction: anisotropic feature deformation}

Refraction describes the change of propagation direction when light crosses an interface between two media with different refractive indices. In geometric optics, the incident and transmitted directions are related by Snell's law,
\begin{equation}
    n_1 \sin \alpha_1 = n_2 \sin \alpha_2,
\end{equation}
where $n_1,n_2$ are the refractive indices of the two media, and $\alpha_1,\alpha_2$ are the angles measured with respect to the interface normal. The essential implication is that a change of medium induces an anisotropic modification of the propagation state: the component associated with the interface normal is altered, while the tangential component is constrained by the transmission condition.

Motivated by this principle, LiNO interprets refraction as a local anisotropic deformation of the latent feature vector. Instead of explicitly solving a geometric-optics transmission problem in the physical space, we introduce an adaptive feature-space normal and use it to separate the latent state into tangential and normal components. Using the same form of adaptive direction as in \eqref{eq:adaptive-normal}, but with independent trainable parameters, we define
\begin{equation}
    h_i=h_i^{\parallel}+h_i^{\perp},
    \qquad
    h_i^{\perp}=\inner{h_i}{\widehat v_i}\widehat v_i,
    \qquad
    h_i^{\parallel}=h_i-h_i^{\perp}.
\end{equation}
The refraction operator is then defined by
\begin{equation}
    \cT_\theta(h_i)
    =
    h_i^{\parallel}+\eta h_i^{\perp},
    \label{eq:refraction}
\end{equation}
where $\eta$ is a learnable refractive factor. 
This construction should be understood as a latent-space analogue inspired by the normal–tangential decomposition underlying refraction, rather than a direct discretization of Snell's law in physical space: the tangential component $h_i^{\parallel}$ is preserved, while the normal component $h_i^{\perp}$ is rescaled according to a learned refractive contrast. Thus, $\eta$ plays the role of an effective refractive-index ratio, controlling how strongly the latent state is bent, stretched, or compressed along the adaptive normal direction.

Equivalently, for a fixed unit vector $\widehat v_i$, \eqref{eq:refraction} can be written as
\begin{equation}
    \cT_\theta(h_i)
    =
    \left(I_M + (\eta-1)\widehat v_i \widehat v_i^\top\right)h_i.
    \label{eq:refraction-matrix}
\end{equation}
Hence, the refraction layer is a rank-one anisotropic deformation of the identity map in the latent feature space. This matrix form makes explicit the connection with a generalized transmission operator: all directions orthogonal to $\widehat v_i$ are left unchanged, whereas the component aligned with $\widehat v_i$ is multiplied by the refractive factor $\eta$.

A naive implementation of \eqref{eq:refraction-matrix} would first construct the dense matrix
\begin{equation}
    T_i = I_M + (\eta-1)\widehat v_i\widehat v_i^\top
    \in \R^{M\times M}
\end{equation}
and then apply it to $h_i$. This would require $O(M^2)$ memory to store $T_i$ and $O(M^2)$ arithmetic operations per spatial point. Such a dense implementation is unnecessary because the refraction operator is only a rank-one perturbation of the identity. Using the normal--tangential decomposition above, we obtain the equivalent matrix-free formula~\eqref{eq:refraction-matrix}.

The matrix-free representation also clarifies the stability role of refraction. For normalized $\widehat v_i$, the local transformation has eigenvalue $\eta$ along the one-dimensional subspace $\operatorname{span}\{\widehat v_i\}$ and eigenvalue $1$ on the orthogonal complement. Consequently, the spectral norm of the local refraction operator is
\begin{equation}
    \norm{T_i}_2 = \max\{1,|\eta|\}.
\end{equation}
Thus, the amplification or attenuation induced by refraction is explicitly controlled by the refractive factor $\eta$.

In the implementation, the refractive factor is constrained near unity by
\begin{equation}
    \eta = 1 + \rho\tanh(\gamma),
    \label{eq:eta-param}
\end{equation}
where $\gamma$ is trainable and $\rho>0$ is a prescribed range parameter. This parameterization has two purposes. First, it initializes the refraction layer close to the identity map, preventing excessive distortion of the latent representation at the beginning of training. Second, it bounds the local amplification and attenuation induced by refraction, which improves the stability of deep compositions of light-evolution blocks. In particular, when $\rho<1$, the factor $\eta$ remains positive and close to one, so the refraction layer acts as a controlled anisotropic modulation rather than an unconstrained feature-space transformation.

Reflection and refraction are complementary. Reflection flips a selected feature component and is approximately norm-preserving when $\widehat v_i$ is normalized, whereas refraction continuously rescales that component. Consequently, reflection provides a discrete geometric reorientation, while refraction provides a smooth anisotropic modulation. Both operations are pointwise in $x$, so they do not replace spatial propagation; rather, they reshape the latent state locally before and after nonlocal scattering. In summary, the refraction layer provides an efficient low-rank feature-space modulation mechanism: it has the expressive effect of an adaptive anisotropic linear map, but its implementation remains matrix-free and linear in the latent width.

\subsection{Scattering as nonlinear spatial kernel propagation}

Scattering is the only component that explicitly couples different spatial locations. Its role is to propagate latent information over the domain through a normalized, input-dependent kernel. We first define query, key, and value embeddings
\begin{equation}
    q_i=W_q h_i,
    \qquad
    k_i=W_k h_i,
    \qquad
    z_i=W_v h_i,
    \label{eq:qkv}
\end{equation}
where $W_q,W_k\in\R^{d\times M}$ and $W_v\in\R^{M\times M}$ are trainable matrices. The standard scattering kernel is
\begin{equation}
    K_{ij}
    =
    \frac{
    \exp\left(d^{-1/2}q_i^\top k_j + b_{ij}\right)
    }{
    \sum_{m=1}^{N}\exp\left(d^{-1/2}q_i^\top k_m + b_{im}\right)
    },
    \label{eq:standard-kernel}
\end{equation}
where $b_{ij}$ is a relative positional bias. For a regular grid, normalized coordinates $p_i\in[0,1]^{d_x}$ are assigned to each spatial location and
\begin{equation}
    b_{ij}=-\tau\norm{p_i-p_j}_2^2,
    \qquad \tau=\operatorname{softplus}(\tau_0)>0.
    \label{eq:relative-bias}
\end{equation}
The parameter $\tau$ controls the locality of the scattering kernel: larger values favor short-range interaction, whereas smaller values allow more global mixing.

The standard scattering update is
\begin{equation}
    (\cS_\theta h)_i
    =
    \sigma_s\left(\sum_{j=1}^{N}K_{ij}z_j-h_i\right),
    \qquad
    \sigma_s=\exp(\lambda_s)>0.
    \label{eq:standard-scattering}
\end{equation}
Because $K_{ij}$ is row-stochastic,
\begin{equation}
    \sum_{j=1}^{N}K_{ij}=1,
\end{equation}
\eqref{eq:standard-scattering} can be interpreted as a relaxation between the current state and a kernel-weighted spatial average. It resembles graph diffusion, collision--scattering relaxation, and nonlocal integral propagation, but with a kernel that depends on the latent field. In the continuum limit, the formal analogue is
\begin{equation}
    (\cS_\theta h)(x)
    =
    \sigma_s\left(
    \int_{\Omega}K_\theta(x,x^\prime;h)Z_\theta h(x^\prime)\,\diff x^\prime - h(x)
    \right),
    \label{eq:continuum-scattering}
\end{equation}
where $Z_\theta$ denotes the value projection and $K_\theta$ is a nonlinear normalized kernel.

\begin{algorithm}[t]
\caption{Standard scattering layer with relative positional bias}
\label{alg:standard-scattering}
\KwIn{Latent field $h\in\R^{B\times N\times M}$; normalized coordinates $p_i\in[0,1]^{d_x}$}
\KwOut{Scattering residual $S\in\R^{B\times N\times M}$}
Compute $q=W_qh$, $k=W_kh$, $z=W_vh$\;
Compute pairwise distances $D_{ij}=\norm{p_i-p_j}_2^2$\;
Set $A_{ij}=d^{-1/2}q_i^\top k_j-\operatorname{softplus}(\tau_0)D_{ij}$\;
Normalize $K_{ij}=\softmax_j(A_{ij})$\;
Compute $y_i=\sum_{j=1}^{N}K_{ij}z_j$\;
Return $S_i=\exp(\lambda_s)(y_i-h_i)$\;
\end{algorithm}

The standard implementation in Algorithm~\ref{alg:standard-scattering} requires forming the dense matrix $K\in\R^{N\times N}$. Therefore, its memory and computational cost scale as $\mathcal{O}(N^2)$ up to channel factors. This is acceptable for moderate grids, but it becomes a bottleneck for high-resolution PDE simulation.

\subsection{Efficient scattering: linearized global kernel plus local diffusion}\label{sec:efficient-scattering}

To improve scalability, we replace the full softmax scattering kernel by a normalized positive separable kernel. The standard softmax kernel relies on the exponential pairwise interaction $\exp(q_i^\top k_j)$ and therefore requires the construction of an $N\times N$ interaction matrix. As detailed in Appendix~\ref{app:exp-kernel-feature}, the exponential dot-product kernel admits an implicit infinite-dimensional feature representation,
\begin{equation}
    \exp(q^\top k)
    =
    \left\langle \Phi(q),\Phi(k)\right\rangle .
\end{equation}
This observation motivates replacing the implicit infinite-dimensional feature map by a finite-dimensional positive feature map 
$\phi:\R^d\to\R_+^d$ and defining the separable kernel surrogate
\begin{equation}\label{eq:positive-separable-kernel}
    \widetilde\kappa(q_i,k_j) = \phi(q_i)^\top\phi(k_j).
\end{equation}
After row normalization, this kernel yields nonnegative scattering weights and preserves the weighted-averaging structure of softmax attention, while avoiding explicit pairwise kernel construction. 
In the implementation, we use
\begin{equation}
    \phi(z)=\ELU(z)+1,
    \label{eq:elu-feature-map}
\end{equation}
which is positive, inexpensive to evaluate, and numerically stable \cite{katharopoulos2020transformers}. 
We emphasize that this deterministic feature map is used as a positive-kernel surrogate rather than an exact finite-dimensional approximation of the exponential kernel.

Because the explicit pairwise positional bias in \eqref{eq:relative-bias} would reintroduce an $N \times N$ matrix, efficient scattering injects spatial information through a coordinate encoder instead. Let
\begin{equation}
    c_i = C_\theta(p_i)\in\R^d
\end{equation}
be a learnable coordinate embedding implemented by a small multilayer perceptron (MLP). The efficient queries and keys are
\begin{equation}
    \widetilde q_i=d^{-1/2}W_qh_i+c_i,
    \qquad
    \widetilde k_i=d^{-1/2}W_kh_i+c_i.
    \label{eq:coord-qk}
\end{equation}
The global linear attention output is then
\begin{equation}
    y_i^{\mathrm{glob}}
    =
    \frac{
    \phi(\widetilde q_i)^\top
    \left(\sum_{j=1}^{N}\phi(\widetilde k_j)z_j^\top\right)
    }{
    \phi(\widetilde q_i)^\top
    \left(\sum_{j=1}^{N}\phi(\widetilde k_j)\right)
    }.
    \label{eq:linear-attention}
\end{equation}
This formulation avoids constructing pairwise interactions and has cost $\mathcal{O}(NdM)$ for a single sample.

The linearized global kernel captures long-range communication but does not explicitly retain the Gaussian locality prior in \eqref{eq:relative-bias}. We therefore add a local diffusion branch
\begin{equation}
    y^{\mathrm{loc}} = P_{\mathrm{loc}}\left(D_{\mathrm{dw}}h\right),
    \label{eq:local-branch}
\end{equation}
where $D_{\mathrm{dw}}$ is a depthwise convolution over the physical grid and $P_{\mathrm{loc}}$ is a pointwise linear projection. In one dimension $D_{\mathrm{dw}}$ is a depthwise one-dimensional convolution; in two dimensions it is a depthwise two-dimensional convolution. The branch is a lightweight surrogate for local scattering and diffusion.

The final efficient scattering average combines global and local contributions:
\begin{equation}
    y_i=(1-\beta)y_i^{\mathrm{glob}}+\beta y_i^{\mathrm{loc}},
    \qquad
    \beta=\operatorname{sigmoid}(\gamma_{\mathrm{loc}})\in(0,1),
    \label{eq:hybrid-scattering-output}
\end{equation}
followed by the scattering residual
\begin{equation}
    (\cS_\theta h)_i=\sigma_s(y_i-h_i).
    \label{eq:efficient-scattering}
\end{equation}
This decomposition approximates the original nonlocal kernel by
\begin{equation}
    \cK_\theta
    \approx
    \cK_\theta^{\mathrm{glob}}+\cK_\theta^{\mathrm{loc}},
    \label{eq:kernel-decomp}
\end{equation}
where $\cK_\theta^{\mathrm{glob}}$ is a low-rank global propagation operator and $\cK_\theta^{\mathrm{loc}}$ is a local diffusion-like operator. The above efficient scattering procedure is summarized in Algorithm~\ref{alg:efficient-scattering}.

\begin{algorithm}[t]
\caption{Efficient scattering layer}
\label{alg:efficient-scattering}
\KwIn{Latent field $h\in\R^{B\times N\times M}$; normalized coordinates $p_i\in[0,1]^{d_x}$}
\KwOut{Efficient scattering residual $S\in\R^{B\times N\times M}$}
Compute $z=W_vh$ and coordinate embeddings $c_i=C_\theta(p_i)$\;
Set $\widetilde q_i=d^{-1/2}W_qh_i+c_i$ and $\widetilde k_i=d^{-1/2}W_kh_i+c_i$\;
Compute positive features $Q_i=\ELU(\widetilde q_i)+1$, $K_i=\ELU(\widetilde k_i)+1$\;
Aggregate $A=\sum_{j=1}^{N}K_jz_j^\top$ and $s_K=\sum_{j=1}^{N}K_j$\;
Compute $y_i^{\mathrm{glob}}=(Q_i^\top A)/(Q_i^\top s_K)$\;
Compute local branch $y^{\mathrm{loc}}=P_{\mathrm{loc}}(D_{\mathrm{dw}}h)$\;
Mix $y_i=(1-\beta)y_i^{\mathrm{glob}}+\beta y_i^{\mathrm{loc}}$, where $\beta=\operatorname{sigmoid}(\gamma_{\mathrm{loc}})$\;
Return $S_i=\exp(\lambda_s)(y_i-h_i)$\;
\end{algorithm}

The efficient scattering layer reduces the dominant spatial complexity from quadratic to linear in $N$:
\begin{equation}
    \mathrm{standard:}\quad \mathcal{O}(N^2(d+M)),
    \qquad
    \mathrm{efficient:}\quad \mathcal{O}(NdM)+\mathcal{O}(NeM),
\end{equation}
where $e$ is the local convolution stencil size. For high-resolution grids, this is the difference between explicitly modeling all pairwise spatial interactions and summarizing them through global kernel statistics plus a local propagation correction.

\subsection{Adaptive mixture of light components}

The relative importance of reflection, refraction, and scattering should depend on the input instance and on the latent state. For example, a coefficient field with sharp inclusions may benefit from stronger local feature reorientation, while a smooth global flow field may require stronger nonlocal propagation. LiNO therefore uses adaptive mixture weights.

Let
\begin{equation}
    \overline h
    =
    \frac{1}{N}\sum_{i=1}^{N}h_i\in\R^M
    \label{eq:global-pool}
\end{equation}
be the spatially averaged latent state. A small gating network $g_\theta:\R^M\to\R^3$ produces
\begin{equation}
    (\alpha_r,\alpha_t,\alpha_s)
    =
    \softmax(g_\theta(\overline h)).
    \label{eq:gating}
\end{equation}
Thus
\begin{equation}
    \alpha_r+\alpha_t+\alpha_s=1,
    \qquad
    \alpha_r,\alpha_t,\alpha_s\ge 0.
\end{equation}
The branch mixture is
\begin{equation}
    m_i
    =
    \alpha_r\cR_\theta(h_i)
    +
    \alpha_t\cT_\theta(h_i)
    +
    \alpha_s(\cS_\theta h)_i.
    \label{eq:branch-mixture}
\end{equation}
The full block then applies a linear residual projection and a pointwise feed-forward correction:
\begin{align}
    \widetilde h_i &= h_i + W_o m_i,
    \label{eq:out-proj}\\
    h_i^+ &= \widetilde h_i + \operatorname{MLP}_\theta(\operatorname{LN}(\widetilde h_i)).
    \label{eq:ffn-correction}
\end{align}
Layer normalization is applied before the light branches and before the feed-forward network. This choice is consistent with stable residual architectures and improves the conditioning of the learned latent evolution.

\section{Results}
\label{sec:experiments}
\subsection{Training objective}

Given training pairs $\{(a^{(n)},u^{(n)})\}_{n=1}^{N_{\mathrm{train}}}$, LiNO is trained by minimizing a composite field-matching objective that combines the mean-square error and the relative $L^2$ error. For stationary or single-shot operator learning tasks, such as coefficient-to-solution maps, we use
\begin{equation}
    \mathcal{J}(\theta)
    =
    \frac{1}{N_{\mathrm{train}}}
    \sum_{n=1}^{N_{\mathrm{train}}}
    \left[
    \norm{\mathcal{G}_{\theta}(a^{(n)})-u^{(n)}}_2
    +
    \frac{
    \norm{\mathcal{G}_{\theta}(a^{(n)})-u^{(n)}}_2
    }{
    \norm{u^{(n)}}_2
    }
    \right]
    \label{eq:training-loss}
\end{equation}
in the current implementation. 
The MSE term penalizes pointwise field discrepancies and provides stable gradient information, while the relative $L^2$ term measures the scale-normalized prediction error commonly reported in neural-operator benchmarks.

For autoregressive time-dependent prediction, we additionally include step-wise rollout errors. Let $\widehat u_{t_m:t_{m+s}}$ and $u_{t_m:t_{m+s}}$ denote the predicted and reference solution segments over the $m$-th output window. The autoregressive relative loss is
\begin{equation}
    \mathcal{J}_{\mathrm{ar}}(\theta)
    =
    \sum_{m=0}^{K-1}
    \frac{
    \norm{\widehat u_{t_m:t_{m+s}}-u_{t_m:t_{m+s}}}_2
    }{
    \norm{u_{t_m:t_{m+s}}}_2
    },
    \label{eq:ar-relative-loss}
\end{equation}
and the full rollout objective is
\begin{equation}
    \mathcal{J}_{\mathrm{rollout}}(\theta)
    =
    \norm{\widehat u_{1:T}-u_{1:T}}_2
    +
    \mathcal{J}_{\mathrm{ar}}(\theta).
    \label{eq:rollout-loss}
\end{equation}

\subsection{Optimization and evaluation}

All models are trained using the composite objectives described above and evaluated by the validation relative $L^2$ error. Unless otherwise stated, we report the best validation performance over training. Detailed implementation choices, optimizer settings, learning-rate schedules, and benchmark-specific hyperparameters are provided in Appendix~\ref{app:implementation-settings}.

\subsection{Burgers' equation}

We first evaluate LiNO on the one-dimensional viscous Burgers benchmark, where the goal is to learn the solution operator from an input function, interpreted as the initial condition, to the scalar solution field on a uniform grid. 
All data are downsampled to resolution $s=256$. 
The input contains the physical field and the normalized coordinate channel, and the output is the scalar solution field.

For LiNO, the full pairwise scattering variant achieves a best validation relative $L^2$ error of $3.19\times10^{-3}$, with a per-epoch training time of approximately $1$ min $54$ s. 
The efficient scattering variant obtains an error of $1.09\times10^{-2}$, while reducing the per-epoch training time to approximately $28$ s. 
Thus, at this moderate one-dimensional resolution, explicit pairwise scattering remains affordable and gives the highest accuracy, whereas efficient scattering provides a favorable accuracy--cost trade-off and is more suitable for higher-dimensional settings where the quadratic kernel becomes expensive.

Table~\ref{tab:burgers-comparison} compares LiNO with representative neural-operator baselines at the same resolution. 
The reference baselines are taken from the FNO benchmark~\cite{li2021fno}, while the DeepONet and LiNO results are obtained using our current implementation and training protocol. 
LiNO with full pairwise scattering gives the lowest error, and the efficient variant also outperforms the reported FNO baseline. 
These results show that the proposed light-evolution block is already competitive on a nonlinear one-dimensional operator-learning task, while retaining a modular scattering mechanism that can be switched between accuracy-oriented and efficiency-oriented regimes.

\begin{table}[t]
\centering
\caption{
Comparison of validation relative $L^2$ errors on the one-dimensional Burgers benchmark at resolution $s=256$.
The published baselines are taken from the FNO benchmark~\cite{li2021fno}; DeepONet and LiNO are trained under our implementation protocol, with detailed settings reported in Appendix~\ref{app:implementation-settings}.
The best LiNO result is highlighted in bold, and the LiNO rows are shaded.
}
\label{tab:burgers-comparison}
\begin{tabular}{lcc}
\toprule
Method & Operator type & Relative $L^2$ error ($\downarrow$)\\
\midrule
NN    & pointwise neural network & $4.71\times10^{-1}$ \\
GCN   & graph convolutional operator & $4.00\times10^{-1}$ \\
FCN   & fully convolutional network & $9.58\times10^{-2}$ \\
PCANN & PCA-based neural operator & $3.98\times10^{-2}$ \\
GNO   & graph kernel neural operator & $5.55\times10^{-2}$ \\
DeepONet & branch--trunk neural operator & $3.50\times10^{-2}$ \\
LNO   & low-rank neural operator & $2.12\times10^{-2}$ \\
MGNO  & multipole graph neural operator & $2.43\times10^{-2}$ \\
FNO   & Fourier neural operator & $1.49\times10^{-2}$ \\
\midrule
\rowcolor{gray!12}
LiNO & efficient scattering & $1.09\times10^{-2}$ \\
\rowcolor{gray!12}
LiNO & full pairwise scattering & $\mathbf{3.19\times10^{-3}}$ \\
\bottomrule
\end{tabular}
\end{table}

\begin{figure}[htbp]
\centering
\includegraphics[width=\linewidth]{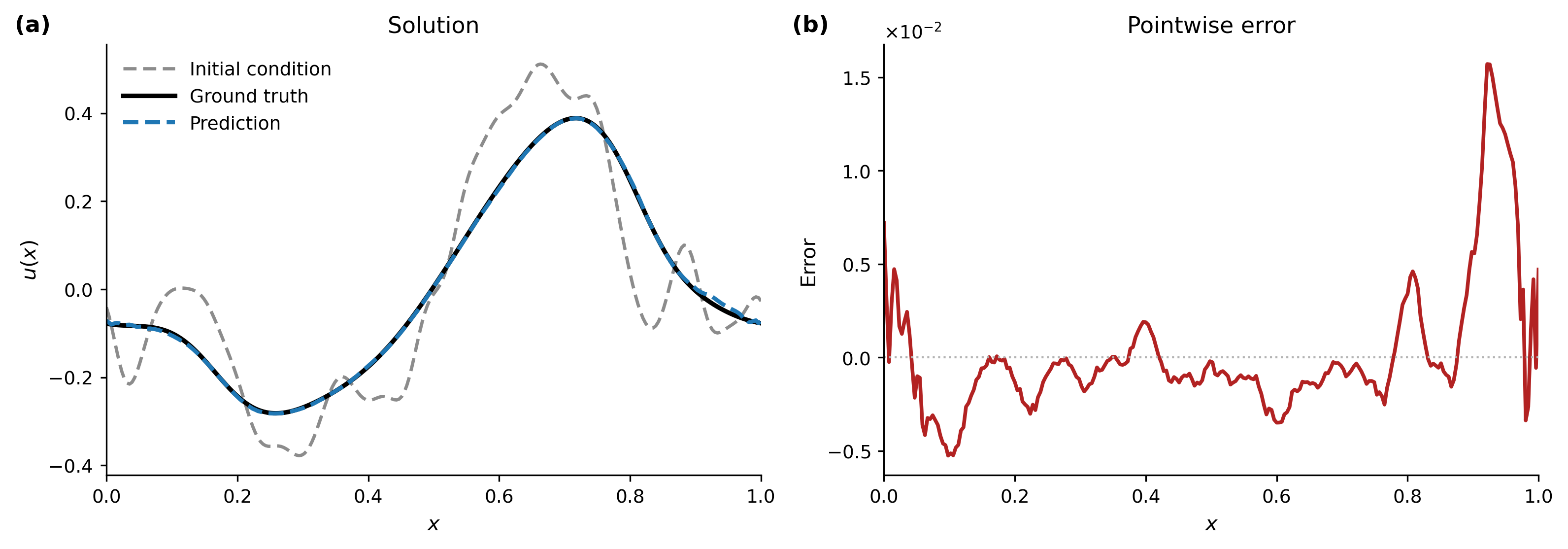}
\caption{
Qualitative result for the one-dimensional viscous Burgers benchmark at resolution $s=256$.
The figure shows the input initial condition, reference solution, LiNO prediction, and pointwise prediction error for a representative validation sample.
}
\label{fig:burgers-qualitative}
\end{figure}

Fig.~\ref{fig:burgers-qualitative} shows a representative validation sample. 
The LiNO prediction closely matches the reference solution over the full spatial domain, including regions with relatively sharp variation. 
The pointwise error remains localized and small, consistent with the relative $L^2$ errors in Table~\ref{tab:burgers-comparison}.

\subsection{Darcy flow}

Next we consider the two-dimensional Darcy flow benchmark, a standard coefficient-to-solution operator-learning problem for elliptic PDEs. 
The input is a heterogeneous coefficient field, and the output is the corresponding scalar solution. 
Compared with Burgers' equation, this problem more directly tests spatial information propagation, since the elliptic solution depends globally on the coefficient field.

We first compare the full pairwise scattering layer and the efficient scattering layer at resolution $85\times85$, keeping all other LiNO components unchanged. 
Both variants achieve nearly identical accuracy: the full pairwise model obtains a validation relative $L^2$ error of $5.94\times10^{-3}$, while the efficient model obtains $5.91\times10^{-3}$. 
Their computational profiles, however, differ substantially. 
The full pairwise scattering layer peaks at about 12.7 GB of allocated CUDA memory and takes roughly 16.5 hours to train, whereas the efficient scattering layer reduces the peak memory to about 3.23 GB and completes training in approximately 4 hours.
Thus, on two-dimensional grids, efficient scattering preserves the accuracy of global spatial propagation while substantially improving memory and runtime scalability.

Table~\ref{tab:darcy-resolution-comparison} compares efficient-scattering LiNO with representative published baselines from the FNO benchmark~\cite{li2021fno} across resolutions $s=85,141,211,421$. 
LiNO consistently outperforms the reported FNO baseline at all tested resolutions. 
The validation relative $L^2$ error increases mildly from $5.91\times10^{-3}$ at $s=85$ to $8.28\times10^{-3}$ at $s=421$, while remaining below the FNO errors across the full resolution range. 
The associated training times are approximately $4$, $22$, $38.5$, and $56$ hours for $s=85,141,211,421$, respectively. 
This confirms that the efficient scattering layer remains trainable at high spatial resolutions where the full pairwise kernel would become increasingly expensive.

\begin{table}[t]
\centering
\caption{
Comparison of validation relative $L^2$ errors on the Darcy flow benchmark at resolutions $s=85,141,211,421$.
LiNO uses efficient scattering and is trained under our implementation and training protocol.
The LiNO row is shaded, and the best result at each resolution is highlighted in bold.
}
\label{tab:darcy-resolution-comparison}
\begin{tabular}{lcccc}
\toprule
Method & $s=85$ & $s=141$ & $s=211$ & $s=421$ \\
\midrule
FCN   & $2.53\times10^{-2}$ & $4.93\times10^{-2}$ & $7.27\times10^{-2}$ & $1.10\times10^{-1}$ \\
PCANN & $2.99\times10^{-2}$ & $2.98\times10^{-2}$ & $2.98\times10^{-2}$ & $2.99\times10^{-2}$ \\
RBM   & $2.44\times10^{-2}$ & $2.51\times10^{-2}$ & $2.55\times10^{-2}$ & $2.59\times10^{-2}$ \\
GNO   & $3.46\times10^{-2}$ & $3.32\times10^{-2}$ & $3.42\times10^{-2}$ & $3.69\times10^{-2}$ \\
LNO   & $5.20\times10^{-2}$ & $4.61\times10^{-2}$ & $4.45\times10^{-2}$ & -- \\
MGNO  & $4.16\times10^{-2}$ & $4.28\times10^{-2}$ & $4.28\times10^{-2}$ & $4.20\times10^{-2}$ \\
FNO   & $1.08\times10^{-2}$ & $1.09\times10^{-2}$ & $1.09\times10^{-2}$ & $9.80\times10^{-3}$ \\
\midrule
\rowcolor{gray!12}
LiNO & $\mathbf{5.91\times10^{-3}}$ & $\mathbf{6.37\times10^{-3}}$ & $\mathbf{6.85\times10^{-3}}$ & $\mathbf{8.28\times10^{-3}}$ \\
\bottomrule
\end{tabular}
\end{table}

We further examine the roles of the three light-evolution components on the Darcy benchmark at resolution $s=85$. 
Table~\ref{tab:darcy2d_ablation_s85} reports the validation relative $L^2$ error obtained by removing one component at a time. 
The scattering component is indispensable: ablating it degrades performance substantially, raising the error from $5.91\times10^{-3}$ to $5.95\times10^{-2}$---nearly an order of magnitude above that of the full-component efficient scattering model.
This confirms that nonlocal spatial propagation is essential for the elliptic coefficient-to-solution map. 
Removing reflection also leads to a measurable degradation, increasing the error to $6.31\times10^{-3}$, whereas removing refraction has only a marginal effect in this example. 
In particular, scattering provides the dominant global communication mechanism, reflection improves the local feature reorientation near coefficient interfaces, and refraction acts as a weaker anisotropic modulation in this benchmark.

\begin{table}[t]
    \centering
    \caption{
    Ablation study of the reflection, refraction, and scattering components on the Darcy2D benchmark at resolution $s=85$.
    Validation relative $L^2$ errors are reported.
    }
    \label{tab:darcy2d_ablation_s85}
    \begin{tabular}{lc}
        \toprule
        Model & Relative $L^2$ error ($\downarrow$) \\
        \midrule
        LiNO w/o scattering  & $5.95 \times 10^{-2}$ \\
        LiNO w/o reflection  & $6.31 \times 10^{-3}$ \\
        LiNO w/o refraction  & $5.94 \times 10^{-3}$ \\
        \midrule
        \rowcolor{gray!12}
        LiNO                 & $\mathbf{5.91 \times 10^{-3}}$ \\
        \bottomrule
    \end{tabular}
\end{table}

\begin{figure}[htbp]
    \centering

    \begin{subfigure}[t]{0.23\textwidth}
        \centering
        \includegraphics[width=\linewidth]{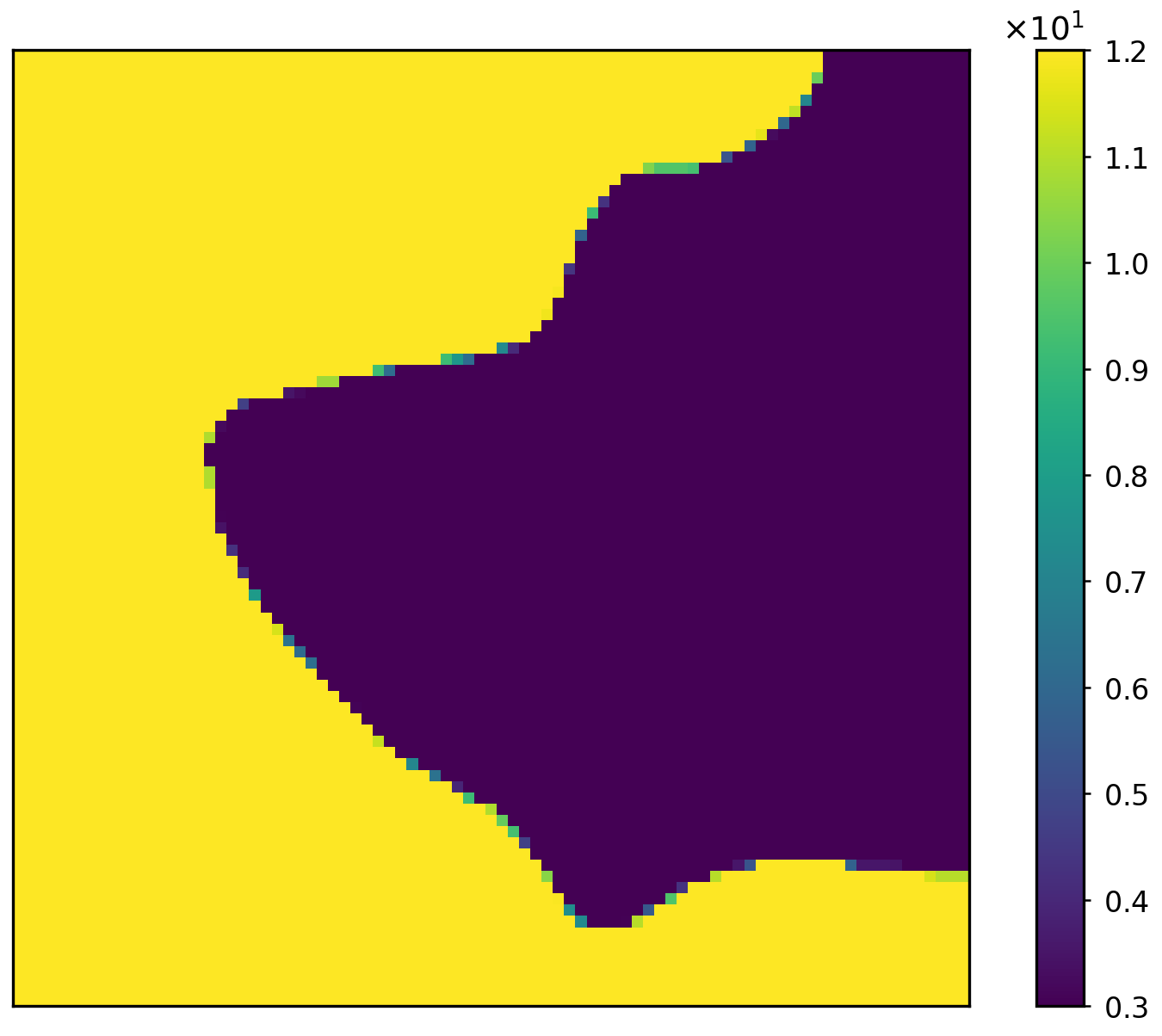}
        \caption*{Coefficient}
        \label{fig:darcy_ablation_coeff}
    \end{subfigure}
    \hspace{0.06\textwidth}
    \begin{subfigure}[t]{0.23\textwidth}
        \centering
        \includegraphics[width=\linewidth]{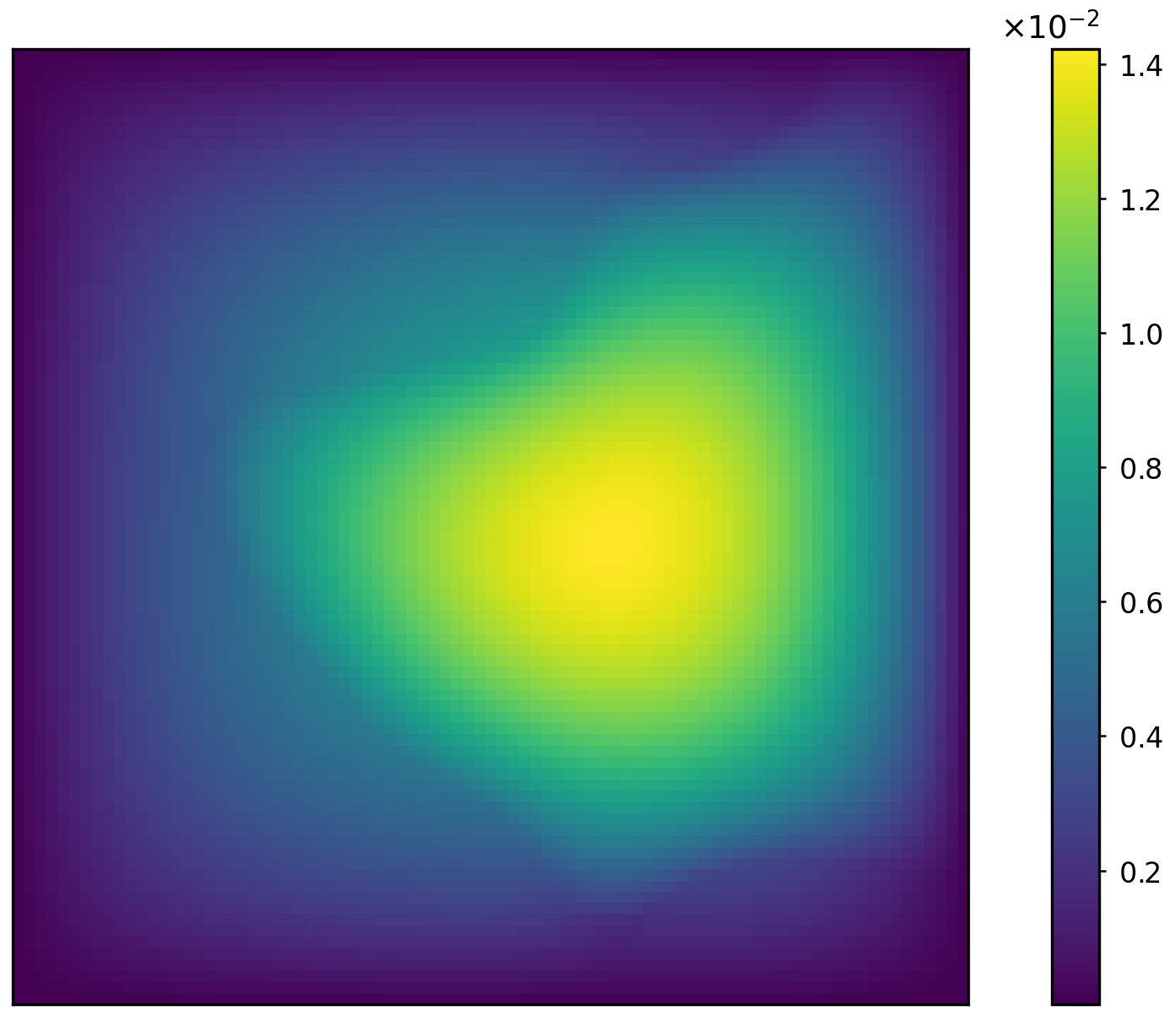}
        \caption*{Ground truth}
        \label{fig:darcy_ablation_gt}
    \end{subfigure}

    \vspace{0.8em}

    \begin{subfigure}[t]{0.23\textwidth}
        \centering
        \includegraphics[width=\linewidth]{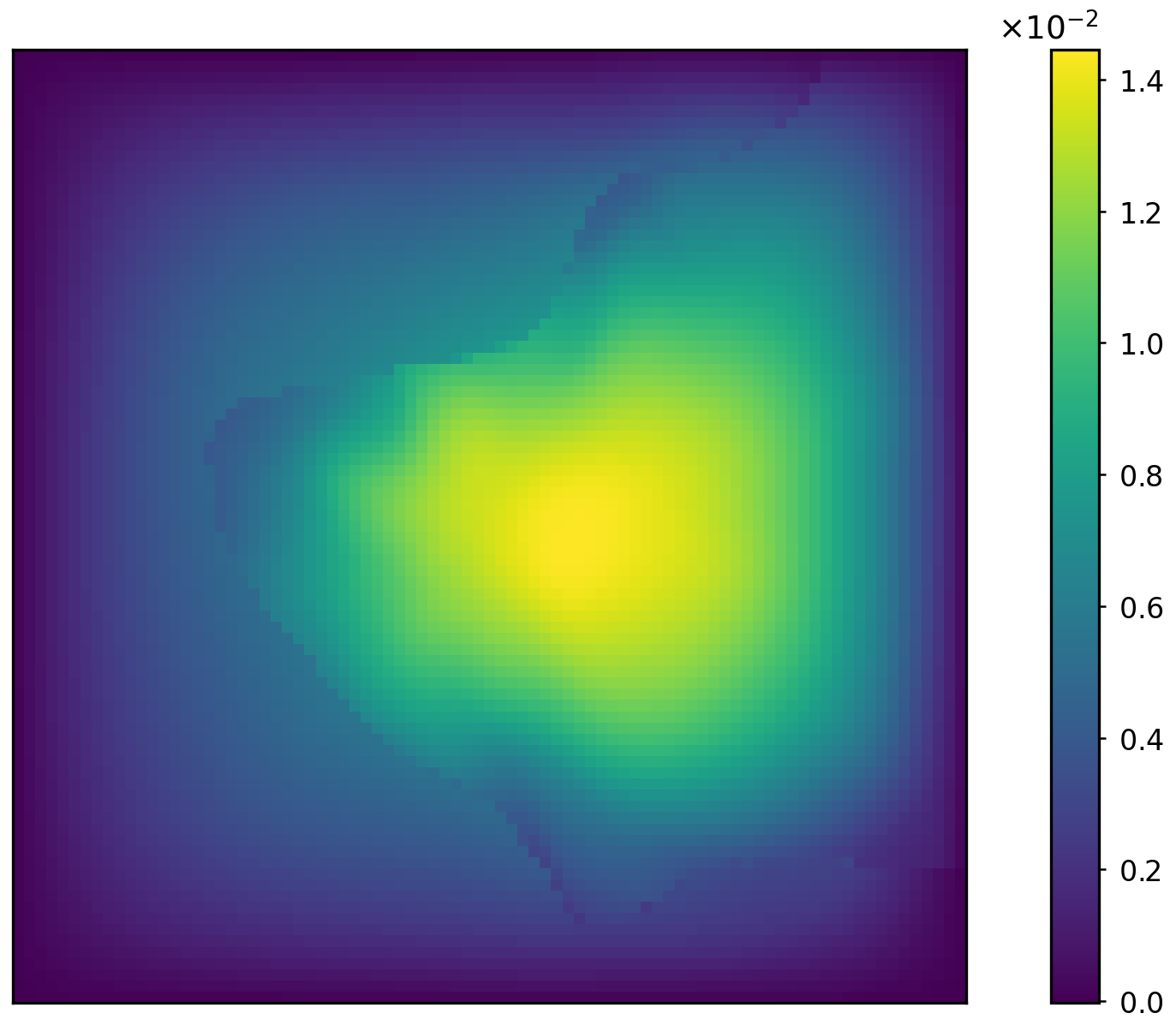}
        \caption*{Pred., w/o scattering}
        \label{fig:darcy_ablation_pred_no_scattering}
    \end{subfigure}
    \hfill
    \begin{subfigure}[t]{0.23\textwidth}
        \centering
        \includegraphics[width=\linewidth]{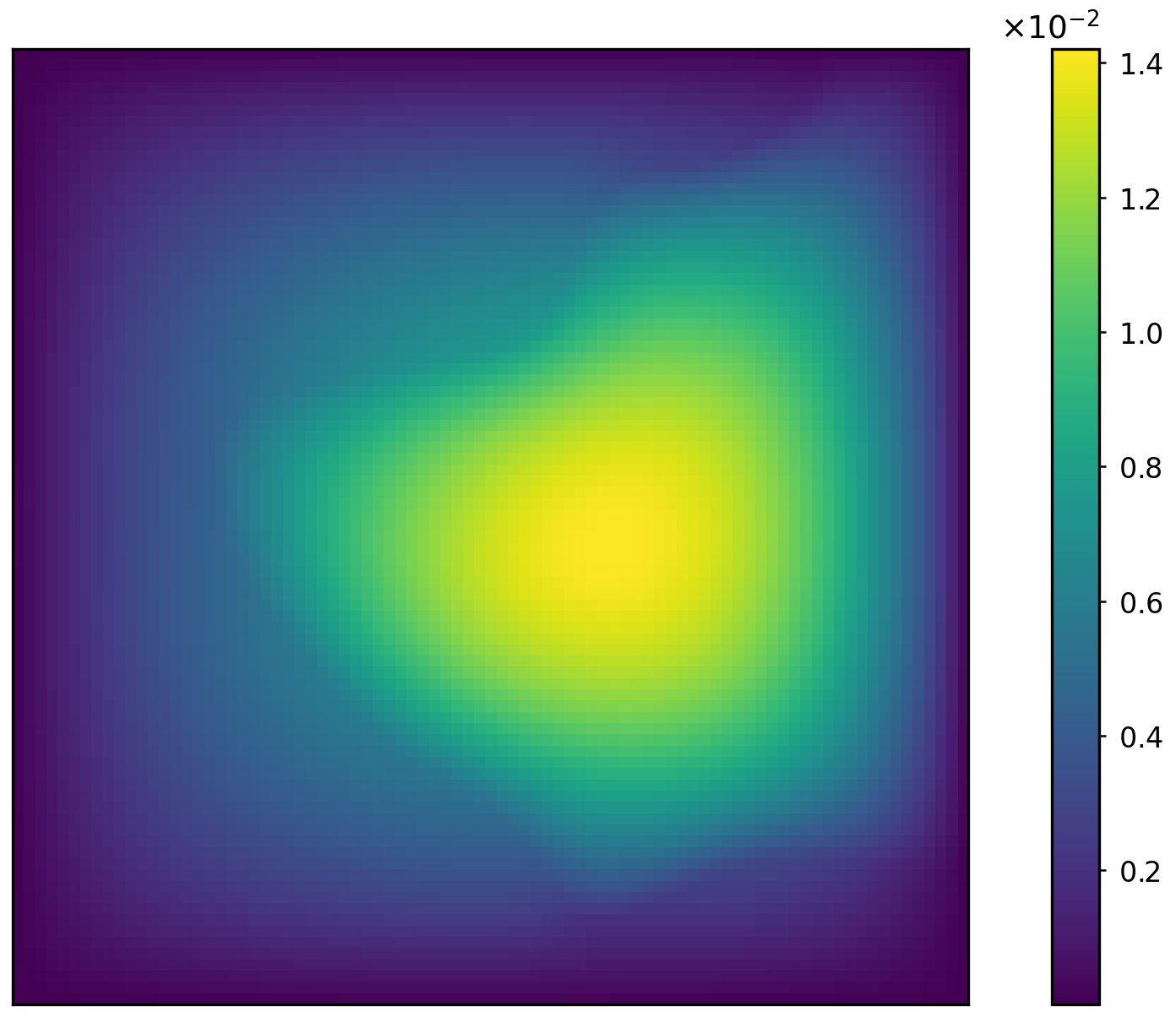}
        \caption*{Pred., w/o reflection}
        \label{fig:darcy_ablation_pred_no_reflection}
    \end{subfigure}
    \hfill
    \begin{subfigure}[t]{0.23\textwidth}
        \centering
        \includegraphics[width=\linewidth]{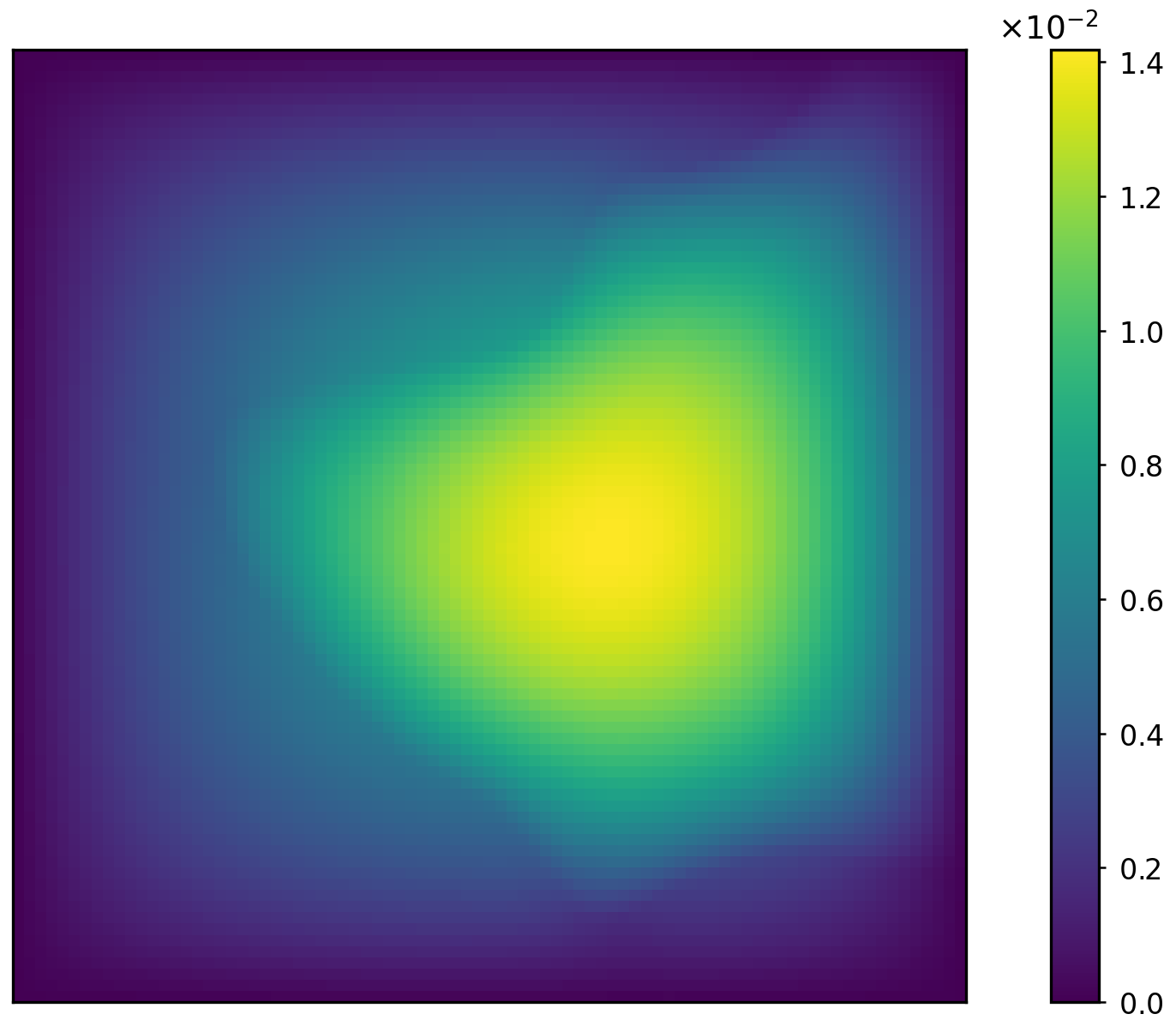}
        \caption*{Pred., w/o refraction}
        \label{fig:darcy_ablation_pred_no_refraction}
    \end{subfigure}
    \hfill
    \begin{subfigure}[t]{0.23\textwidth}
        \centering
        \includegraphics[width=\linewidth]{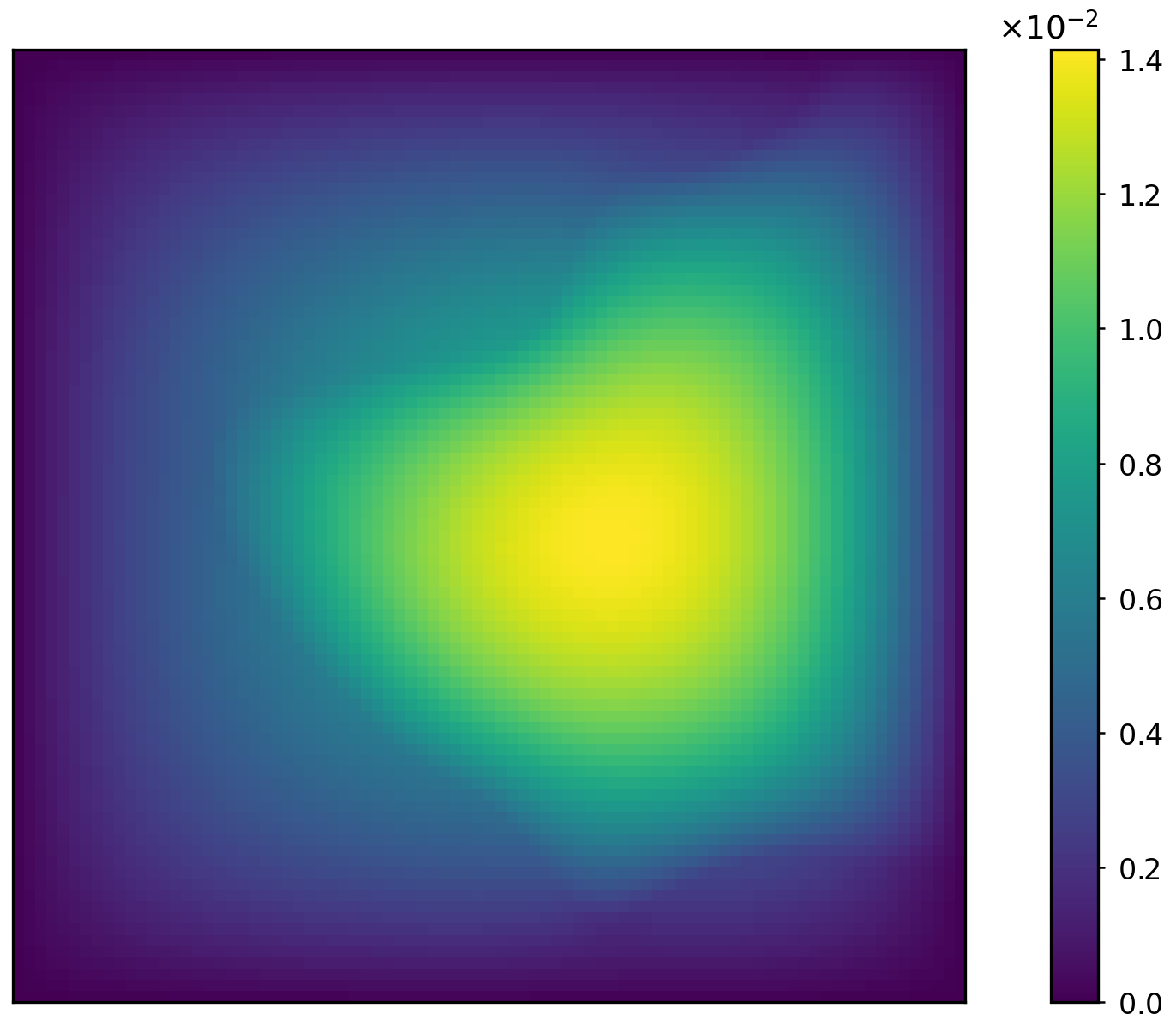}
        \caption*{Pred., full model}
        \label{fig:darcy_ablation_pred_full}
    \end{subfigure}

    \vspace{0.8em}

    \begin{subfigure}[t]{0.23\textwidth}
        \centering
        \includegraphics[width=\linewidth]{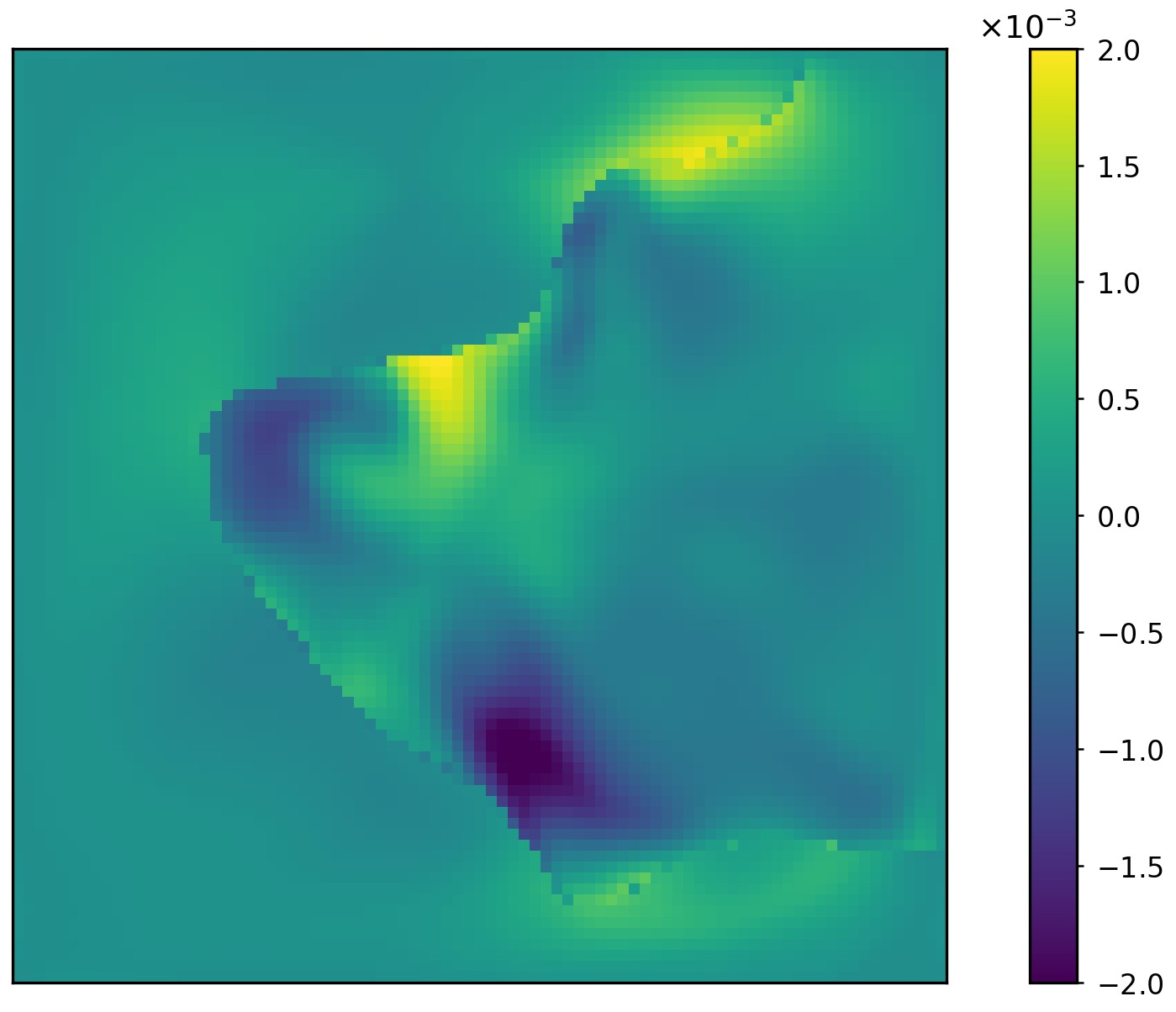}
        \caption*{Error, w/o scattering}
        \label{fig:darcy_ablation_err_no_scattering}
    \end{subfigure}
    \hfill
    \begin{subfigure}[t]{0.23\textwidth}
        \centering
        \includegraphics[width=\linewidth]{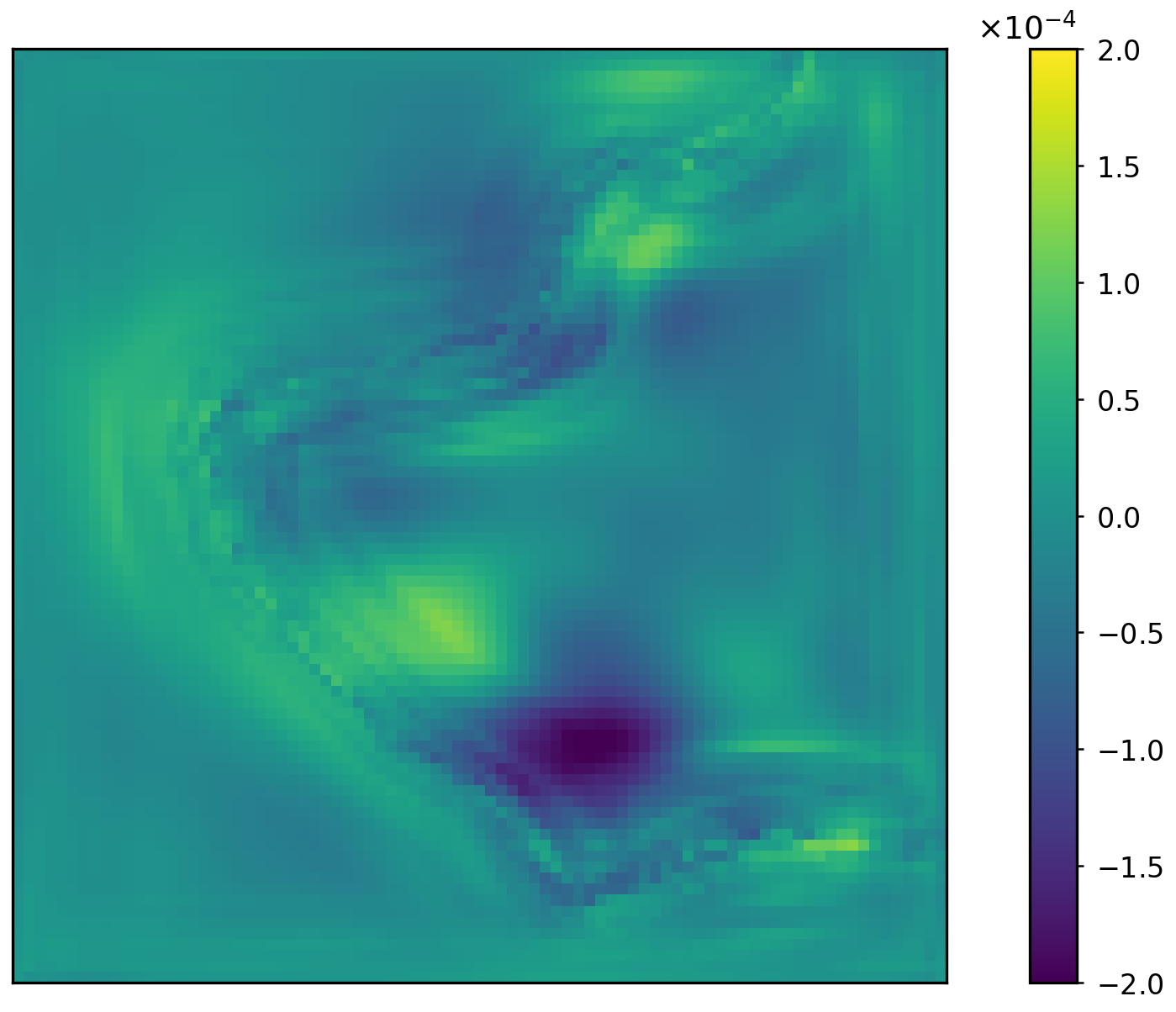}
        \caption*{Error, w/o reflection}
        \label{fig:darcy_ablation_err_no_reflection}
    \end{subfigure}
    \hfill
    \begin{subfigure}[t]{0.23\textwidth}
        \centering
        \includegraphics[width=\linewidth]{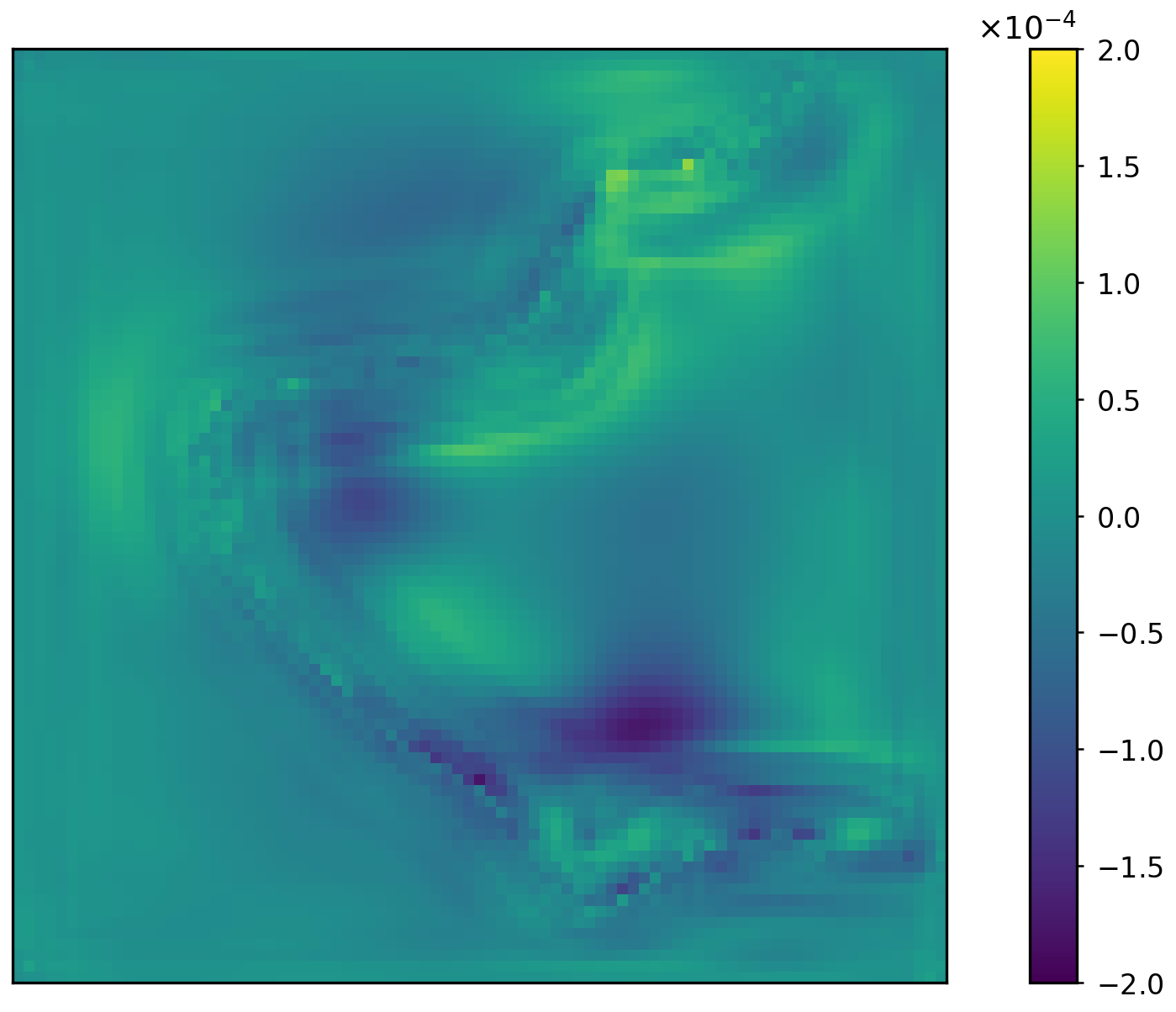}
        \caption*{Error, w/o refraction}
        \label{fig:darcy_ablation_err_no_refraction}
    \end{subfigure}
    \hfill
    \begin{subfigure}[t]{0.23\textwidth}
        \centering
        \includegraphics[width=\linewidth]{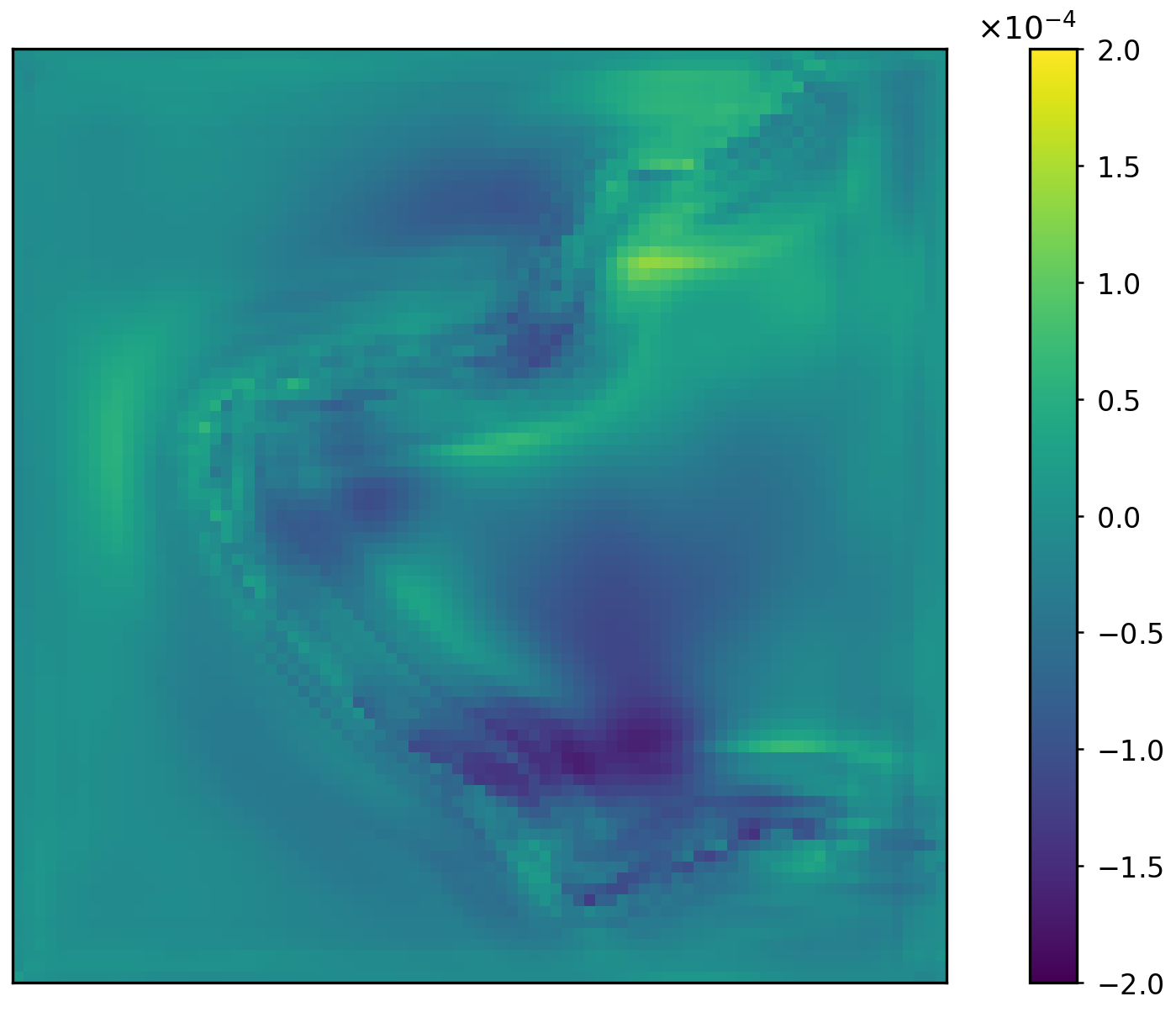}
        \caption*{Error, full model}
        \label{fig:darcy_ablation_err_full}
    \end{subfigure}

    \caption{
    Qualitative ablation study on the Darcy benchmark at resolution $s=85$.
    The first row shows the input coefficient field and the reference solution.
    The second row shows predictions obtained without scattering, without reflection, without refraction, and with the full model.
    The third row shows the corresponding pointwise errors.
    Removing scattering causes the largest global degradation, while removing reflection mainly affects the reconstruction near coefficient interfaces.
    }
    \label{fig:darcy2d_ablation_s85}
\end{figure}

Fig.~\ref{fig:darcy2d_ablation_s85} provides a qualitative comparison for the same ablation setting. 
The model without scattering produces visibly larger errors over the whole domain. 
In particular, the predicted solution shows distorted level-set structures and spurious interface-aligned artifacts near the discontinuous coefficient region. 
This is consistent with the elliptic nature of Darcy flow: the solution at each location is affected by the global arrangement of the permeability field, so removing scattering weakens the nonlocal communication needed to transmit information across different parts of the domain. 
As a result, the model can no longer correctly balance the global pressure distribution induced by the heterogeneous coefficient interface.
The model without reflection remains close to the full model in the global solution profile, but its error is more concentrated around the coefficient interface. 
This suggests that reflection mainly contributes to local feature reorientation near sharp material transitions. 
From the optical viewpoint, the coefficient interface acts like a change of medium: latent features crossing this interface need to be redirected so that the model can represent the change in solution gradients across regions of different permeability. 
Without the reflection branch, the prediction still captures the overall elliptic response through scattering, but the local transition near the interface becomes less sharp and less accurately aligned with the underlying coefficient geometry.
By contrast, removing refraction causes only minor visual changes in this example, consistent with its small quantitative effect in Table~\ref{tab:darcy2d_ablation_s85}. 
This indicates that, for this Darcy setting, the dominant difficulty is the combination of global elliptic coupling and interface-sensitive local redirection, rather than smooth anisotropic rescaling of latent features. 
Overall, the ablation results show that scattering is indispensable for global spatial coupling, reflection provides an interface-sensitive correction for sharp coefficient transitions, and refraction plays a secondary role for this benchmark.

\subsection{Airfoil flow}

\begin{figure}[htbp]
\centering
\includegraphics[width=0.9\linewidth]{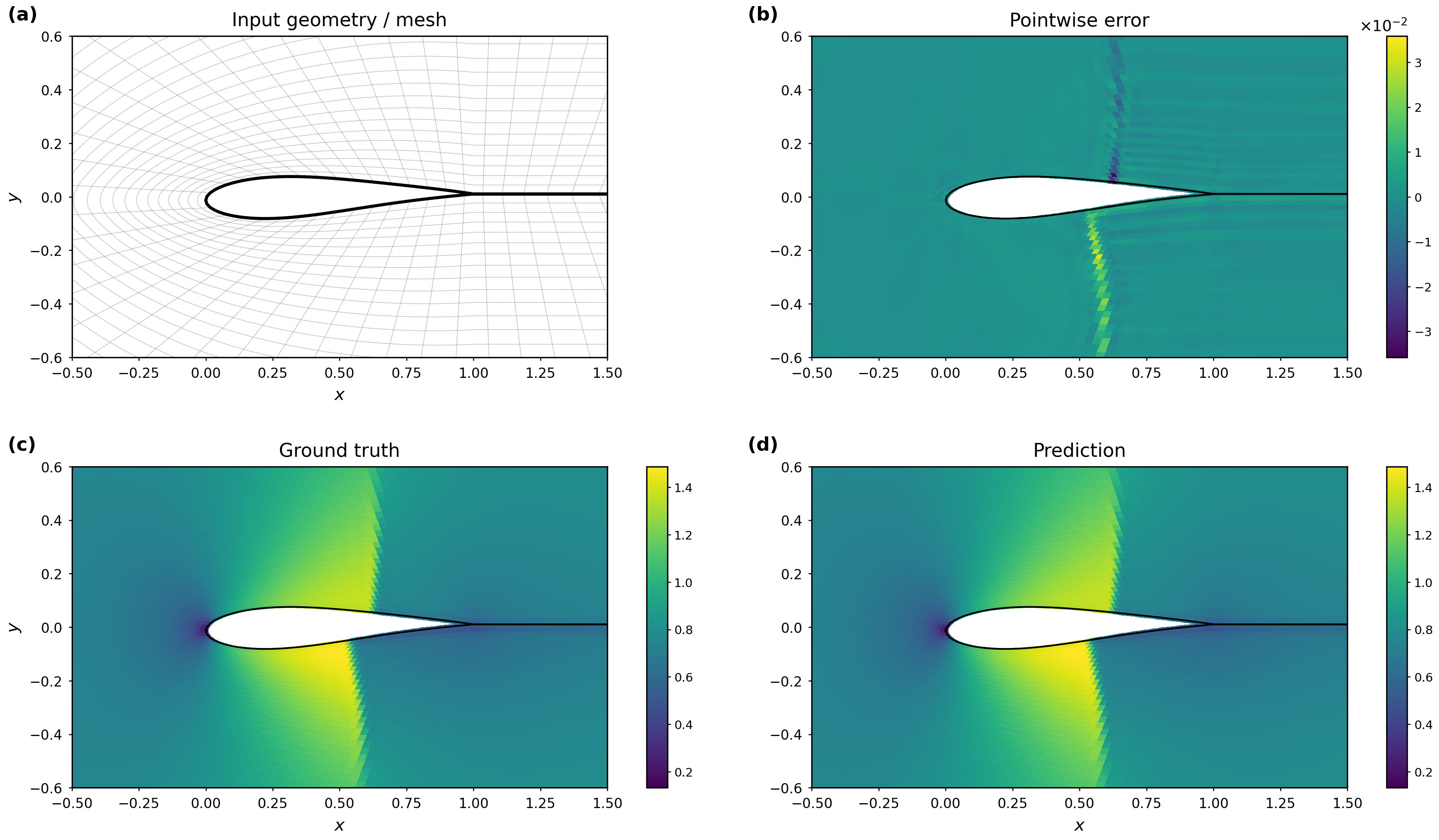}
\caption{
Qualitative evaluation on the transonic airfoil benchmark.
Panel (a) shows the input body-fitted geometry and mesh, panel (b) shows the pointwise prediction error, and panels (c)--(d) compare the reference and predicted Mach number fields.
LiNO captures the global flow pattern and the near-airfoil structures, while the largest errors remain localized near regions with strong spatial variation.
}
\label{fig:airfoil-qualitative}
\end{figure}

Next we evaluate LiNO on a geometry-dependent transonic airfoil benchmark. 
Unlike the Burgers and Darcy examples, this problem is defined on a body-fitted curvilinear mesh and therefore tests whether the proposed light-evolution block can handle spatially deformed computational domains represented through mesh coordinates. 
The input consists of the mesh point locations, and the target is the Mach number field on the same mesh. 
Fig.~\ref{fig:airfoil-qualitative} shows a representative validation example. 
LiNO accurately reproduces the global Mach distribution around the deformed airfoil and captures the main near-body flow structures. 
The prediction agrees well with the reference solution in both the far-field region and the vicinity of the airfoil surface. 
The pointwise error remains spatially localized, with larger discrepancies mainly appearing near regions of strong flow variation around the airfoil. 
This behavior is consistent with the increased difficulty of geometry-dependent operator learning, where the model must learn both the dependence on the deformed mesh and the associated flow response.

These results indicate that the same LiNO architecture used for Cartesian-grid Burgers and Darcy problems can also be applied to body-fitted airfoil data. 
Although the current implementation treats the curvilinear mesh as a structured grid with coordinate channels, the experiment suggests that the light-inspired latent evolution remains effective beyond simple rectangular domains. 
A natural next step is to replace the grid-based scattering layer with mesh-, graph-, or point-cloud-based scattering for fully unstructured aerodynamic geometries.

\subsection{Two-dimensional Navier--Stokes dynamics}

Finally, we evaluate LiNO on autoregressive prediction of two-dimensional Navier--Stokes vorticity dynamics. 
Unlike the stationary Burgers and Darcy benchmarks, this task requires repeatedly applying the learned one-step operator in time. 
It therefore tests whether the efficient-scattering LiNO can maintain stable short-horizon predictions under error accumulation.

\begin{figure}[htbp]
\centering
\includegraphics[width=\linewidth]{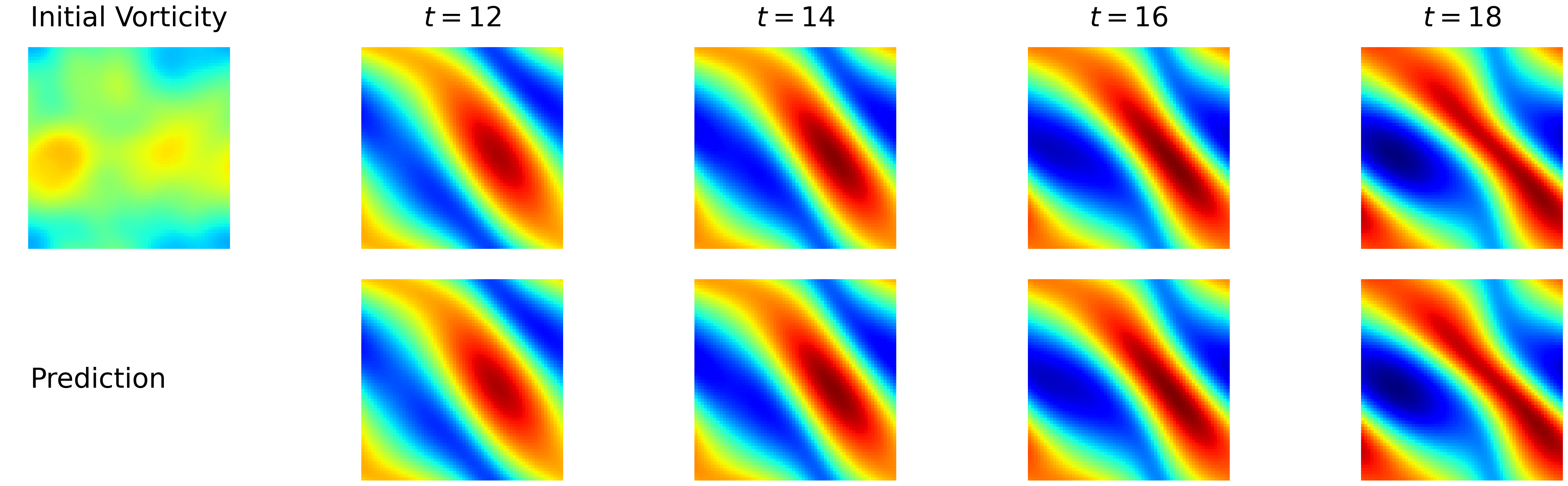}
\caption{
Autoregressive prediction on the two-dimensional Navier--Stokes benchmark at resolution $s = 64$.
The left panel shows the initial vorticity field.
The remaining columns compare reference vorticity fields in the upper row and LiNO predictions in the lower row at selected rollout times.
}
\label{fig:ns-qualitative}
\end{figure}

The vorticity trajectories are sampled on a $64\times64$ periodic grid. 
Following the standard autoregressive neural-operator setting, the first $T_{\mathrm{in}}=10$ vorticity frames are used as the input history. 
At each rollout step, LiNO takes the most recent ten observed or predicted vorticity fields, concatenated with two normalized coordinate channels, and predicts the next vorticity field. 
Thus, the input tensor has shape $B\times64\times64\times12$, and the one-step output has shape $B\times64\times64\times1$. 
The model is rolled out for $T_{\mathrm{out}}=10$ future steps. 
We use $2000$ trajectories for training and $1000$ trajectories for validation. The model and optimizer settings follow Table~\ref{tab:app-training-settings}.
The best checkpoint reaches a one-step relative error of $2.80\times10^{-3}$ and a full-rollout relative error of $2.92\times10^{-3}$. 
At the final epoch, the corresponding errors are $3.17\times10^{-3}$ and $3.35\times10^{-3}$. 
The small gap between the one-step and rollout errors indicates that the learned dynamics remain stable under autoregressive composition over the ten-step prediction horizon.
Fig.~\ref{fig:ns-qualitative} shows a representative validation trajectory. 
The reference vorticity fields are displayed in the upper row and the LiNO predictions in the lower row. 
LiNO accurately preserves the dominant coherent structures, their phase locations, and the large-scale transport patterns throughout the rollout. 
Small discrepancies appear mainly in fine-scale regions, but no rapid drift or qualitative instability is observed.

\begin{figure}[htbp]
\centering
\includegraphics[width=0.5\linewidth]{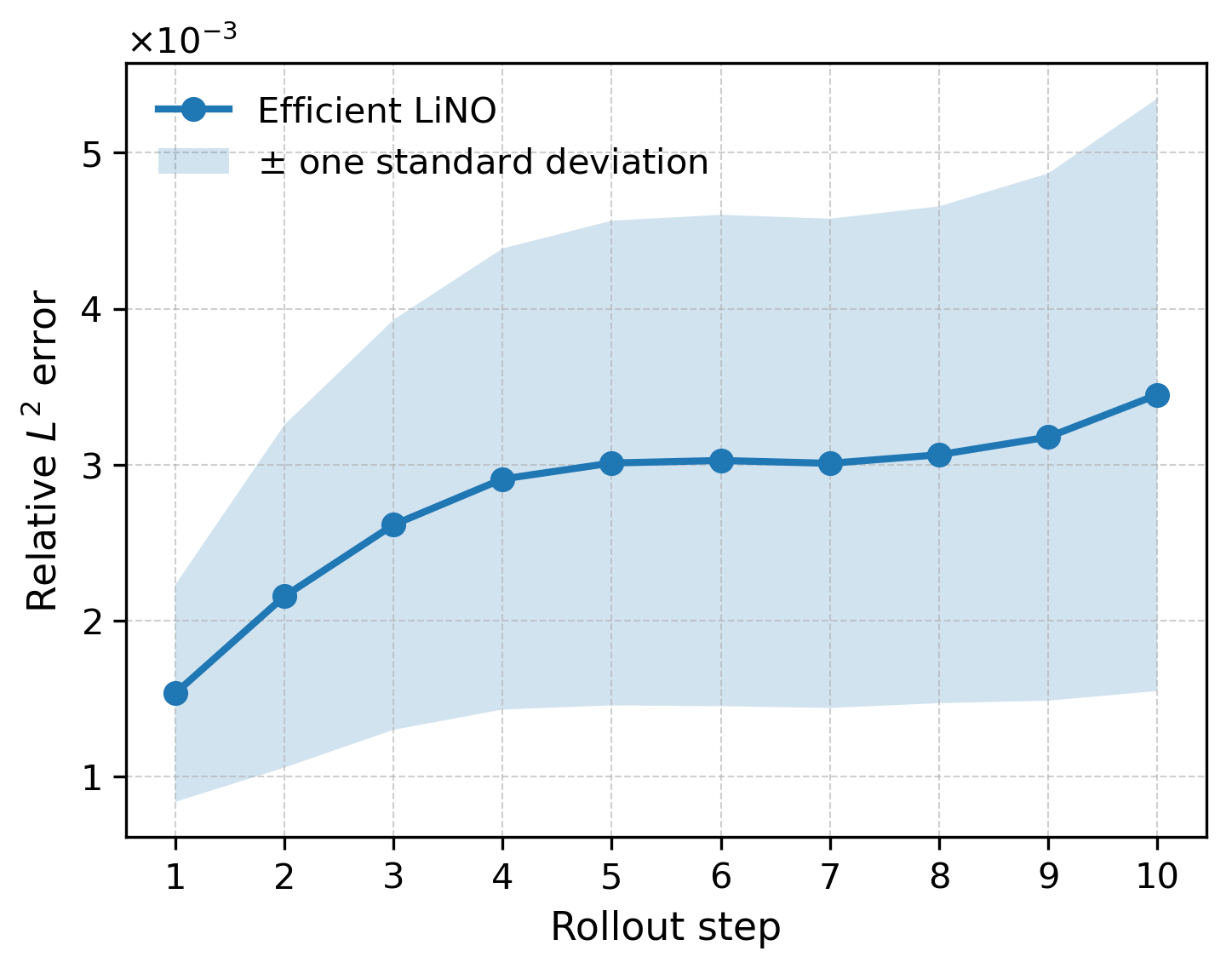}
\caption{
Temporal error behavior of efficient-scattering LiNO on the two-dimensional Navier--Stokes benchmark.
The curve reports the mean step-wise relative $L^2$ error over the autoregressive rollout horizon, and the shaded region denotes one standard deviation across validation trajectories.
}
\label{fig:ns-temporal-error}
\end{figure}

To quantify temporal stability, we report the step-wise relative error along the rollout horizon. 
For the $k$-th future step, the error is computed by averaging the relative $L^2$ discrepancy over all validation trajectories,
\begin{equation}
\mathrm{RelErr}_k
=
\frac{1}{N_{\rm val}}
\sum_{n=1}^{N_{\rm val}}
\frac{
\left\|
\widehat{\omega}_{t_{\rm in}+k}^{(n)}
-
\omega_{t_{\rm in}+k}^{(n)}
\right\|_{2}
}{
\left\|
\omega_{t_{\rm in}+k}^{(n)}
\right\|_{2}
},
\qquad
k=1,\ldots,T_{\rm out}.
\label{eq:temporal-relerr}
\end{equation}
Fig.~\ref{fig:ns-temporal-error} shows that the mean error increases gradually from approximately $1.5\times10^{-3}$ at the first prediction step to about $3.5\times10^{-3}$ at the final step. 
The error remains below $4\times10^{-3}$ throughout the rollout, and the standard-deviation band shows moderate sample-to-sample variation without unstable growth. 
These results indicate that efficient scattering provides sufficient nonlocal spatial communication for stable learned time marching in this vorticity benchmark.

\section{Conclusion}
\label{sec:conclusion}

We have introduced the Light-inspired Neural Operator (LiNO), a neural-operator architecture that organizes latent evolution through three interpretable mechanisms inspired by light transport: reflection, refraction, and scattering. 
Reflection and refraction act as adaptive pointwise transformations in the latent feature space, enabling local feature reorientation and anisotropic modulation. 
Scattering provides nonlocal spatial propagation through a learned kernel over the physical domain. 
This decomposition separates local feature transformation from global information exchange, yielding a modular operator-learning framework that is compatible with both one- and two-dimensional PDE surrogate modeling.
A key component of LiNO is its scattering mechanism. 
The full pairwise scattering layer provides an expressive normalized kernel with relative positional bias, while the efficient scattering layer replaces the quadratic pairwise interaction by positive-feature linearized global propagation combined with a local diffusion branch. 
This design preserves the role of nonlocal spatial communication while substantially reducing memory and runtime costs on two-dimensional grids. 
The ablation study on Darcy flow further confirms that scattering is the dominant component for elliptic coefficient-to-solution maps, while reflection improves local feature reconstruction near coefficient interfaces and refraction provides an additional but weaker anisotropic modulation.

Numerical experiments demonstrate that the same light-evolution block can be used across several PDE operator-learning settings. 
On the one-dimensional Burgers benchmark, full pairwise scattering achieves high accuracy when exact global interactions remain affordable. 
On Darcy flow, efficient-scattering LiNO consistently improves over the reported FNO baseline across multiple resolutions, while remaining trainable up to $s=421$. 
On the geometry-dependent transonic airfoil benchmark, LiNO captures the global Mach field and near-body flow structures on a body-fitted curvilinear mesh. 
On two-dimensional Navier--Stokes vorticity dynamics, efficient-scattering LiNO maintains stable autoregressive rollout, with step-wise errors remaining below $4\times10^{-3}$ over the ten-step prediction horizon. 
Together, these results indicate that light-inspired latent transport provides a useful inductive bias for building interpretable and scalable neural operators.

Several directions remain open. 
First, the current implementation mainly treats structured Cartesian or curvilinear grids. 
Extending LiNO to fully unstructured meshes, point clouds, and moving-boundary geometries would broaden its applicability to realistic engineering simulations. 
This requires replacing grid-based scattering with mesh-, graph-, or point-cloud-based propagation while preserving the reflection--refraction--scattering decomposition. 
Second, the scattering mechanism can be further generalized to adaptive resolutions, multiscale discretizations, and heterogeneous sampling densities, which are important for localized sharp layers, interfaces, shocks, and turbulent structures. 
Third, incorporating physical constraints such as conservation, boundary conditions, symmetry, equivariance, or energy stability into the light-evolution block could improve reliability in long-time prediction and high-fidelity simulation. 
Finally, a more systematic theoretical analysis of approximation, stability, and resolution transfer would help clarify when and why light-inspired latent evolution improves operator learning.

Overall, LiNO provides a step toward physically motivated neural operators whose internal mechanisms are not purely black-box transformations, but are structured around interpretable geometric and transport principles. 
We expect this perspective to be useful for developing scalable neural surrogates for complex-domain PDEs, geometry-dependent physical systems, and large-scale scientific computing applications.

\section*{Data availability}
The Burgers, Darcy flow, and Navier–Stokes benchmark datasets used in this study are publicly accessible from the Fourier Neural Operator benchmark repository, while the transonic airfoil benchmark data are available through the Point Cloud Neural Operator (PCNO) model data repository.

\section*{Code availability}
The source code is available at \url{https://github.com/wukekever/Light-inspired-neural-operator}.

\section*{Acknowledgements}
Jingrun Chen is supported by NSFC Major Research Plan -  Interpretable and General-purpose Next-generation Artificial Intelligence (Nos 92570001 and 92370205), NSFC grant 12425113, and the Key Laboratory of the Ministry of Education for Mathematical Foundations and Applications of Digital Technology, University of Science and Technology of China. Keke Wu is supported by the China Postdoctoral Science Foundation under Grant Number 2025M773105 and the Jiangsu Funding Program for Excellent Postdoctoral Talent. Yixuan Zhang is supported by the NSFC of China (12501582). 
We thank Professor Weinan E for helpful discussions.


\section*{Competing interests}
The authors declare no competing interests.

\appendix

\section{Appendix}
\label{app:benchmark-pdes}

This appendix summarizes the PDE benchmarks used in the numerical experiments. 
For each problem, we specify the underlying equation, the initial and boundary 
conditions, and the operator-learning map approximated by LiNO.

\subsection{One-dimensional Burgers' equation}
\label{app:burgers}

The first benchmark is the one-dimensional viscous Burgers' equation on the unit interval,
\begin{equation}
    \partial_t u(x,t)
    +
    \partial_x\left(\frac{1}{2}u(x,t)^2\right)
    =
    \nu \partial_{xx}u(x,t),
    \qquad
    (x, t)\in (0,1)\times(0,T].
    \label{eq:app-burgers}
\end{equation}
We use the standard Burgers benchmark associated with the FNO dataset, with viscosity $\nu = 0.1$, and $T = 1$.
The initial condition is
\begin{equation}
    u(x,0)=u_0(x),
    \qquad x\in(0,1),
    \label{eq:app-burgers-ic}
\end{equation}
where $u_0$ is sampled from the prescribed random function distribution used in the benchmark dataset. 
Periodic boundary conditions are imposed,
\begin{equation}
    u(0,t)=u(1,t),
    \qquad
    \partial_x u(0,t)=\partial_x u(1,t),
    \qquad t\in[0,T].
    \label{eq:app-burgers-bc}
\end{equation}

The operator-learning task is to approximate the final-time solution map
\begin{equation}
    \mathcal{G}_{\mathrm{Burgers}}:
    u_0 \mapsto u(\cdot,T).
    \label{eq:app-burgers-map}
\end{equation}
In our experiments, the spatial field is downsampled to a resolution of $s = 256$ with $1600$ samples used for training and $448$ for validation.

\subsection{Two-dimensional Darcy flow}
\label{app:darcy}

The second benchmark is a two-dimensional elliptic Darcy flow problem on the unit square
$\Omega=(0,1)^2$. 
Given a heterogeneous permeability coefficient $a(x)$, the pressure or potential field $u(x)$ satisfies
\begin{equation}
    -\nabla\cdot\big(a(x)\nabla u(x)\big)=f(x),
    \qquad x = (x_1, x_2) \in\Omega
    \label{eq:app-darcy}
\end{equation}
with homogeneous Dirichlet boundary condition
\begin{equation}
    u(x)=0,
    \qquad x\in\partial\Omega.
    \label{eq:app-darcy-bc}
\end{equation}
Following the standard Darcy benchmark, the forcing term is fixed as
\begin{equation}
    f(x)=1,
    \qquad x\in\Omega.
    \label{eq:app-darcy-forcing}
\end{equation}
The input coefficient $a(x)$ is a heterogeneous piecewise coefficient field and may contain sharp interfaces. 
The corresponding operator-learning task is to approximate the coefficient-to-solution map
\begin{equation}
    \mathcal{G}_{\mathrm{Darcy}}:
    a \mapsto u.
    \label{eq:app-darcy-map}
\end{equation}
We use $800$ samples for training and $224$ samples for validation. 
We evaluate LiNO at multiple spatial resolutions,
\begin{equation}
    s\in\{85,141,211,421\},
    \label{eq:app-darcy-resolutions}
\end{equation}
corresponding to grids of size $85^2$, $141^2$, $211^2$, and $421^2$. 

\subsection{Transonic airfoil flow}
\label{app:airfoil}

The airfoil benchmark considers inviscid transonic flow over a deformed airfoil. 
The governing equations are the two-dimensional compressible Euler equations,
\begin{align}
    \frac{\partial \rho}{\partial t}
    + \nabla\cdot(\rho \bm{v})
    &= 0,
    \label{eq:app-airfoil-mass}
    \\
    \frac{\partial \rho\bm{v}}{\partial t}
    + \nabla\cdot\left(\rho\bm{v}\otimes\bm{v}+p\I\right)
    &= 0,
    \label{eq:app-airfoil-momentum}
    \\
    \frac{\partial E}{\partial t}
    + \nabla\cdot\left((E+p)\bm{v}\right)
    &= 0,
    \label{eq:app-airfoil-energy}
\end{align}
where $\rho$ is the density, $\bm{v}$ is the velocity, $p$ is the pressure, and $E$ is the total energy. 
Viscous effects are neglected. 
The far-field condition is specified by
\begin{equation}
    \rho_{\infty}=1,\qquad
    p_{\infty}=1.0,\qquad
    M_{\infty}=0.8,\qquad
    \mathrm{AoA}=0,
    \label{eq:app-airfoil-farfield}
\end{equation}
where $M_{\infty}$ is the free-stream Mach number and $\mathrm{AoA}$ is the angle of attack. 
A no-penetration boundary condition is imposed on the airfoil surface.

The airfoil geometry is parameterized by a design-element deformation. 
The initial NACA-0012 profile is mapped to a cubic design element with eight control nodes. 
New airfoil shapes are generated by vertically displacing these control nodes, with displacement sampled from $\mathbb{U}[-0.05,0.05]$.
The resulting deformed airfoil is discretized using a body-fitted mesh.
The operator-learning task is to approximate the map from the mesh geometry to the corresponding Mach number field,
\begin{equation}
    \mathcal{G}_{\mathrm{airfoil}}:
    (x_1, x_2) \mapsto M(x_1, x_2),
    \label{eq:app-airfoil-map}
\end{equation}
where $(x_1, x_2)$ denotes the physical mesh coordinates and $M(x_1, x_2)$ is the Mach number evaluated at the mesh points. 
In the implementation, the input tensor contains the physical coordinate channels of the body-fitted mesh, and the output tensor contains the scalar Mach number field.

\subsection{Two-dimensional incompressible Navier--Stokes equation}
\label{app:navier-stokes}

The fourth benchmark is the two-dimensional incompressible Navier--Stokes equation in vorticity form on the periodic domain $\Omega=(0,1)^2$. 
The vorticity field $\omega(x,t)$ satisfies
\begin{equation}
    \partial_t \omega(x,t)
    +
    v(x,t)\cdot\nabla \omega(x,t)
    =
    \nu \Delta \omega(x,t)
    +
    f(x),
    \qquad
    (x,t)\in \Omega\times(0,T],
    \label{eq:app-ns-vorticity}
\end{equation}
where $v=(v_1,v_2)$ is the incompressible velocity field, $\nu>0$ is the viscosity, and $f$ is a prescribed time-independent forcing term. We use $\nu = 10^{-3}$ and the forcing term
\begin{equation}
    f(x_1,x_2)
    =
    0.1\left[
    \sin\big(2\pi(x_1+x_2)\big)
    +
    \cos\big(2\pi(x_1+x_2)\big)
    \right].
    \label{eq:app-ns-forcing}
\end{equation}
The velocity field is divergence-free,
\begin{equation}
    \nabla\cdot v(x,t)=0,
    \label{eq:app-ns-incompressible}
\end{equation}
and is recovered from the vorticity through the stream function $\psi$,
\begin{equation}
    v = \nabla^\perp \psi
    =
    \big(\partial_{x_2}\psi,-\partial_{x_1}\psi\big),
    \qquad
    -\Delta \psi = \omega.
    \label{eq:app-ns-stream}
\end{equation}
The initial condition is
\begin{equation}
    \omega(x,0)=\omega_0(x),
    \qquad x\in\Omega,
    \label{eq:app-ns-ic}
\end{equation}
where $\omega_0$ is sampled from the prescribed random initial distribution in the benchmark dataset. 
Periodic boundary conditions are imposed in both spatial directions.

The operator-learning task is autoregressive prediction of future vorticity fields. 
Given a temporal history of $T_{\mathrm{in}}=10$ vorticity snapshots,
\begin{equation}
    \big(
    \omega(\cdot,t_{m-T_{\mathrm{in}}+1}),
    \ldots,
    \omega(\cdot,t_m)
    \big),
\end{equation}
the model predicts the next vorticity field $\omega(\cdot,t_{m+1})$. 
The corresponding one-step map is
\begin{equation}
    \mathcal{G}_{\mathrm{NS}}:
    \big(
    \omega_{m-T_{\mathrm{in}}+1},
    \ldots,
    \omega_m
    \big)
    \mapsto
    \omega_{m+1}.
    \label{eq:app-ns-map}
\end{equation}
Repeated application of $\mathcal{G}_{\mathrm{NS}}$ gives the full autoregressive rollout. 
In our experiments, we predict $T_{\mathrm{out}}=10$ future frames.

The vorticity fields are discretized on a uniform grid with spatial resolution $s = 64$,
corresponding to a grid of size $64\times64$. 
At each autoregressive step, the input tensor consists of the most recent ten vorticity frames concatenated with two normalized coordinate channels.
Thus, each input sample has shape $64\times64\times12$, and each one-step output sample has shape $64\times64\times1$.

We use $2000$ trajectories for training and $1000$ trajectories for validation. 
During evaluation, the model is rolled out autoregressively for ten future steps by feeding each predicted vorticity field back into the temporal history.

\subsection{Implementation and hyperparameter settings}
\label{app:implementation-settings}

We summarize the implementation details and training hyperparameters used in the experiments. 
All datasets are converted to channel-last PyTorch tensors before training. For one-dimensional fields, the input tensor has shape $B\times Q\times d_{\mathrm{in}}$, whereas for two-dimensional fields it has shape $B\times H\times W\times d_{\mathrm{in}}$. The lifting map projects the last dimension to the latent width $M$, the stacked light-evolution blocks preserve the spatial shape and latent width, and the projection map outputs $d_{\mathrm{out}}$ channels. Physical input channels and target fields are normalized by unit Gaussian normalization using statistics computed from the training set. Coordinate channels, when used, are concatenated to the physical input channels and are not normalized.

All experiments are conducted on a single NVIDIA A800 80G GPU with a 48-core Intel Xeon Gold 6342 CPU. 
Unless otherwise stated, all LiNO models use latent width $M=128$ and $L=8$ light-evolution blocks, where each block contains the three proposed branches: reflection, refraction, and scattering. The efficient scattering layer is used as the default scattering module, while the full pairwise scattering layer is retained as a reference implementation for small-resolution comparison and ablation experiments. All models are trained with AdamW using an initial learning rate of $10^{-3}$, weight decay $10^{-5}$, batch size $4$, and gradient clipping with maximum norm $1.0$. The random seed is fixed to $42$ for both data splitting and model initialization.

Benchmark-specific settings are reported in Table~\ref{tab:app-training-settings}. In the learning-rate schedule row, StepLR$(s,\gamma)$ denotes a step learning-rate scheduler with step size $s$ and decay factor $\gamma$. For the Darcy ablation study at resolution $s=85$, we use the same configuration as the efficient-scattering Darcy model and remove one light-evolution component at a time. The DeepONet baseline used in the Burgers comparison is trained under the same data split, normalization, optimizer, batch size, and number of epochs as the Burgers LiNO models; its architecture-specific settings are reported in Table~\ref{tab:app-baseline-settings}.

\begin{table}[htbp]
\centering
\caption{Training hyperparameter settings across benchmarks. Common settings are latent width 
$M=128$, depth $L=8$, batch size $4$, AdamW optimizer, initial learning rate $10^{-3}$, weight decay 
$10^{-5}$, random seed $42$, and gradient clipping with maximum norm $1.0$.}
\label{tab:app-training-settings}
\small
\setlength{\tabcolsep}{6pt}
\begin{tabular}{lcccc}
\toprule
\textbf{Hyperparameter} 
& \textbf{Burgers} 
& \textbf{Darcy} 
& \textbf{Airfoil} 
& \textbf{Navier--Stokes} \\
\midrule
Scattering module 
& full / efficient 
& efficient$^\dagger$ 
& efficient 
& efficient \\
Epochs 
& 500 
& 500 
& 1000 
& 500 \\
LR schedule 
& StepLR$(5,0.96)$ 
& StepLR$(5,0.96)$ 
& StepLR$(200,0.5)$ 
& StepLR$(5,0.96)$ \\
\bottomrule
\end{tabular}

\vspace{0.4em}
\begin{minipage}{0.95\linewidth}
\footnotesize
$^\dagger$ For Darcy, full pairwise scattering is used only in the $85\times85$ efficiency comparison; 
the default Darcy model and all ablation variants use efficient scattering.
\end{minipage}
\end{table}

\begin{table}[htbp]
\centering
\caption{Architecture-specific settings for the DeepONet baseline on the Burgers benchmark.}
\label{tab:app-baseline-settings}
\small
\setlength{\tabcolsep}{8pt}
\begin{tabular}{lc}
\toprule
\textbf{Hyperparameter} & \textbf{DeepONet} \\
\midrule
Branch input & Initial condition $u_0$ \\
Trunk input & Normalized spatial coordinate $x$ \\
Basis dimension & $128$ \\
Branch/Trunk width & $128$ \\
Branch/Trunk depth & $8$ fully connected layers \\
Output channels & $1$ \\
Optimizer & AdamW \\
Initial learning rate & $10^{-3}$ \\
Weight decay & $10^{-5}$ \\
Batch size & $4$ \\
Epochs & $500$ \\
LR schedule & StepLR$(5,0.96)$ \\
\bottomrule
\end{tabular}
\end{table}

\section{Feature representation of the exponential dot-product kernel}
\label{app:exp-kernel-feature}

This appendix is intended only to motivate the use of positive separable kernels for efficient scattering. The specific implementation in the main experiments uses $\phi(z)=\ELU(z)+1$, which should be viewed as a practical 
linear-attention feature map rather than a truncated Taylor or Mercer approximation of the exponential kernel.

For $q,k\in\R^d$, the exponential dot-product kernel admits the Taylor expansion
\begin{equation}
    \exp(q^\top k)
    =
    \sum_{m=0}^{\infty}
    \frac{(q^\top k)^m}{m!}.
    \label{eq:app-exp-taylor}
\end{equation}
For each integer $m\geq 1$, the $m$-th power of the dot product can be written as an inner product between tensor-product features:
\begin{equation}
    (q^\top k)^m
    =
    \left\langle
    q^{\otimes m},
    k^{\otimes m}
    \right\rangle,
    \label{eq:app-tensor-inner}
\end{equation}
where
\begin{equation}
    q^{\otimes m}
    =
    \underbrace{q\otimes q\otimes \cdots \otimes q}_{m\ \mathrm{times}}
\end{equation}
denotes the $m$-fold tensor product of $q$. 
Indeed, since
\begin{equation}
    q^\top k
    =
    \sum_{\alpha=1}^{d}q_\alpha k_\alpha,
\end{equation}
we have
\begin{equation}
    (q^\top k)^m
    =
    \sum_{\alpha_1=1}^{d}\cdots\sum_{\alpha_m=1}^{d}
    q_{\alpha_1}\cdots q_{\alpha_m}
    k_{\alpha_1}\cdots k_{\alpha_m},
\end{equation}
which is precisely the Euclidean inner product between $q^{\otimes m}$ and $k^{\otimes m}$. 
For $m=0$, we use the convention
\begin{equation}
    q^{\otimes 0}=1,
    \qquad
    k^{\otimes 0}=1.
\end{equation}

Combining \eqref{eq:app-exp-taylor} and \eqref{eq:app-tensor-inner}, each Taylor term can be expressed as
\begin{equation}
    \frac{(q^\top k)^m}{m!}
    =
    \left\langle
    \frac{1}{\sqrt{m!}}q^{\otimes m},
    \frac{1}{\sqrt{m!}}k^{\otimes m}
    \right\rangle.
    \label{eq:app-scaled-tensor-inner}
\end{equation}
Therefore, the exponential dot-product kernel has the infinite-dimensional feature representation
\begin{equation}
    \exp(q^\top k)
    =
    \left\langle
    \Phi(q),
    \Phi(k)
    \right\rangle,
    \label{eq:app-exp-feature-inner}
\end{equation}
where the feature map $\Phi$ is given by
\begin{equation}
    \Phi(q)
    =
    \left(
    1,\,
    q,\,
    \frac{1}{\sqrt{2!}}q^{\otimes 2},\,
    \frac{1}{\sqrt{3!}}q^{\otimes 3},\,
    \ldots,\,
    \frac{1}{\sqrt{m!}}q^{\otimes m},\,
    \ldots
    \right).
    \label{eq:app-infinite-feature-map}
\end{equation}

\paragraph{Connection to Mercer's theorem.}
The derivation above is a concrete instance of Mercer's theorem. Mercer’s theorem states that for a continuous symmetric positive-definite kernel $K(x,y)$ defined on a compact set $\mathcal{X}\times\mathcal{X}$, there exists a Hilbert space $\mathcal{H}$ and a feature map $\Phi:\mathcal{X}\to\mathcal{H}$ such that
\begin{equation}
    K(x,y)=\langle\Phi(x),\Phi(y)\rangle_{\mathcal{H}},
\end{equation}
where the inner product is taken in $\mathcal{H}$. This theorem guarantees that any positive-definite kernel can be represented as an inner product in some (possibly infinite-dimensional) feature space.

For the exponential dot-product kernel $K(q,k)=\exp(q^\top k)$, we have explicitly constructed this feature map $\Phi$, whose image space $\mathcal{H}$ is the Hilbert direct sum of tensor spaces:
\begin{equation}
    \mathcal{H}=\bigoplus_{m=0}^{\infty}\mathcal{H}_m,\qquad\text{where }\mathcal{H}_m=(\R^d)^{\otimes m}.
\end{equation}
The corresponding inner product is the sum of inner products at each order:
\begin{equation}
    \langle\Phi(q),\Phi(k)\rangle_{\mathcal{H}}
    =
    \sum_{m=0}^{\infty}
    \left\langle
    \frac{q^{\otimes m}}{\sqrt{m!}},
    \frac{k^{\otimes m}}{\sqrt{m!}}
    \right\rangle_{\mathcal{H}_m}.
\end{equation}
This is precisely the explicit form of the Mercer decomposition: the kernel is decomposed into a countable infinity of orthogonal feature functions (here, tensor monomials) with weights $1/m!$.

Note that Mercer's theorem classically requires the kernel to be defined on a compact domain and to satisfy square-integrability conditions. The exponential dot-product kernel is defined on all of $\R^d$, so strictly speaking one needs additional regularization (e.g., restricting $\|q\|,\|k\|\leq R$) to apply the classical theorem directly. However, as shown by equation~\eqref{eq:app-feature-map-norm}, the feature map has finite norm as long as $\|q\|$ is finite, so the construction remains valid on any locally compact set.

The map $\Phi(q)$ is well defined as an element of the Hilbert direct sum of tensor spaces. Indeed,
\begin{equation}
    \|\Phi(q)\|^2
    =
    \sum_{m=0}^{\infty}
    \frac{\|q^{\otimes m}\|^2}{m!}
    =
    \sum_{m=0}^{\infty}
    \frac{\|q\|^{2m}}{m!}
    =
    \exp(\|q\|^2)
    <
    \infty .
    \label{eq:app-feature-map-norm}
\end{equation}
This expression shows that the exponential kernel implicitly contains polynomial tensor features of all orders.

We next record several special cases that illustrate the structure and computational implications of the infinite-dimensional representation.

\paragraph{One-dimensional case.}
When $d=1$, the vectors $q$ and $k$ reduce to scalars. In this case, $q^\top k=qk$, and the exponential kernel becomes
\begin{equation}
    \exp(qk)
    =
    \sum_{m=0}^{\infty}
    \frac{q^m k^m}{m!}.
\end{equation}
The corresponding feature map is
\begin{equation}
    \Phi(q)
    =
    \left(
    1,\,
    q,\,
    \frac{q^2}{\sqrt{2!}},\,
    \frac{q^3}{\sqrt{3!}},\,
    \ldots,\,
    \frac{q^m}{\sqrt{m!}},\,
    \ldots
    \right).
\end{equation}
Thus, even in one dimension, the exponential kernel contains polynomial features of all degrees.

\paragraph{Two-dimensional case.}
When $d=2$, let
\begin{equation}
    q=(q_1,q_2),
    \qquad
    k=(k_1,k_2).
\end{equation}
The first-order term is
\begin{equation}
    q^\top k=q_1k_1+q_2k_2.
\end{equation}
The second-order term is
\begin{equation}
    (q^\top k)^2
    =
    q_1^2k_1^2
    +
    2q_1q_2k_1k_2
    +
    q_2^2k_2^2.
\end{equation}
Equivalently, it can be written as
\begin{equation}
    (q^\top k)^2
    =
    \left\langle
    \left(q_1^2,\sqrt{2}q_1q_2,q_2^2\right),
    \left(k_1^2,\sqrt{2}k_1k_2,k_2^2\right)
    \right\rangle .
\end{equation}
This is the compressed symmetric-tensor representation of the second-order full tensor product. 
It shows explicitly that the exponential dot-product kernel contains not only linear interactions between $q$ and $k$, but also higher-order polynomial interactions.

\paragraph{Truncated feature expansion and its Mercer perspective.}
If the infinite series is truncated at order $P$, one obtains the finite-dimensional polynomial approximation
\begin{equation}
    \exp(q^\top k)
    \approx
    \sum_{m=0}^{P}
    \frac{(q^\top k)^m}{m!}
    =
    \left\langle
    \Phi_P(q),
    \Phi_P(k)
    \right\rangle,
\end{equation}
where
\begin{equation}
    \Phi_P(q)
    =
    \left(
    1,\,
    q,\,
    \frac{1}{\sqrt{2!}}q^{\otimes 2},\,
    \ldots,\,
    \frac{1}{\sqrt{P!}}q^{\otimes P}
    \right).
\end{equation}

From the Mercer perspective, truncation to order $P$ amounts to retaining only the first $P+1$ Mercer eigencomponents and discarding all higher-order ones:
\begin{equation}
    \exp(q^\top k)\approx\sum_{m=0}^{P}\frac{(q^\top k)^m}{m!}
    =
    \sum_{m=0}^{P}\lambda_m\,\psi_m(q)\,\psi_m(k),
\end{equation}
where $\lambda_m=1/m!$ and $\psi_m(q)=q^{\otimes m}$ (under the appropriate normalization). This reveals the nature of the truncated approximation: it is a \emph{finite-rank Mercer approximation} that approximates the infinite-dimensional kernel with finitely many feature functions. The error arises from the discarded high-order Mercer components:
\begin{equation}
    \text{Error}
    =
    \sum_{m=P+1}^{\infty}\frac{(q^\top k)^m}{m!}
    \leq
    \frac{\exp(|q^\top k|)\cdot|q^\top k|^{P+1}}{(P+1)!}.
\end{equation}
When $|q^\top k|$ is small, the truncation error decays factorially with $P$.

However, its dimension grows rapidly with both $d$ and $P$. 
If the full tensor products are used, the number of features up to order $P$ is
\begin{equation}
    1+d+d^2+\cdots+d^P,
\end{equation}
while using symmetric tensor features reduces the order-$m$ dimension to
\begin{equation}
    \binom{d+m-1}{m}.
\end{equation}
Nevertheless, the total dimension still grows quickly as $P$ increases. 
Therefore, explicitly constructing tensor-product features is generally impractical for high-dimensional attention or scattering layers. 
The efficient scattering layer instead uses a simple positive feature map $\phi$ to construct a computationally tractable separable kernel surrogate.

\paragraph{Zero-query or zero-key case.}
If $q=0$ or $k=0$, then
\begin{equation}
    \exp(q^\top k)=1.
\end{equation}
In the infinite-dimensional representation, all higher-order tensor features vanish, and only the constant feature remains:
\begin{equation}
    \Phi(0)=(1,0,0,\ldots).
\end{equation}
This case illustrates the role of the constant component in the feature map: it provides a nonzero baseline interaction when the dot-product similarity vanishes.

\paragraph{From Mercer decomposition to separable surrogate kernels.}
The Mercer analysis above shows that the exact exponential kernel requires an infinite-dimensional feature representation. In practice, explicitly constructing the tensor features $\Phi(q)$ is infeasible (the dimension grows exponentially with $d$ and $P$). The efficient scattering layer therefore does not attempt to reproduce the infinite Mercer map $\Phi$ exactly. Instead, it introduces a \emph{finite-dimensional, componentwise-positive surrogate feature map} $\phi:\R^d\to\R^D_+$ (with $D\ll\infty$) and constructs a separable surrogate kernel:
\begin{equation}
    \widetilde\kappa(q_i,k_j)
    =
    \phi(q_i)^\top \phi(k_j).
    \label{eq:app-separable-surrogate}
\end{equation}
This can be viewed as a \emph{practical variant} of Mercer's theorem: rather than demanding an exact representation of the original kernel, one seeks a computationally feasible positive-definite kernel that preserves the normalized weighted-averaging structure of softmax attention. From the operator-theoretic viewpoint, this amounts to approximating the original infinite-rank integral operator by a finite-rank positive-definite operator.

After normalization over the index $j$, this surrogate defines nonnegative scattering weights:
\begin{equation}
    \widetilde K_{ij}
    =
    \frac{
    \phi(q_i)^\top\phi(k_j)
    }{
    \sum_{\ell=1}^{N}\phi(q_i)^\top\phi(k_\ell)
    },
    \qquad
    \widetilde K_{ij}\geq 0,
    \qquad
    \sum_{j=1}^{N}\widetilde K_{ij}=1,
\end{equation}
provided that $\phi$ is componentwise positive, as is the case for $\phi(z)=\ELU(z)+1$. 
Thus, the finite-dimensional positive feature map preserves the normalized weighted-averaging structure of softmax attention while allowing the scattering output to be computed without forming an explicit $N\times N$ pairwise kernel matrix. 
In this sense, the feature map used in Section~\ref{sec:efficient-scattering} should be understood as a positive-kernel surrogate rather than an exact finite-dimensional approximation of the exponential kernel.

\bibliographystyle{unsrt} 
\bibliography{references.bib}
\end{document}